# Training with synthetic data for drone detection in thermal imagery

Tanel Liiv*[a], Sander Soodla[a], Nzamba Bignoumba[a], Alma M. Liezenga[b], Toomas Pruuden[a]
[a]Marduk Technologies OÜ, Tallinn, Estonia; [b]Intelligent Imaging, TNO Defence, Safety and Security, The Hague, The Netherlands

## ABSTRACT

Ground-to-Air (G2A) drone detection in the long-wave infrared (LWIR) spectral bands presents distinct challenges, including reduced texture information, sensor noise, and weak thermal contrast. The development of robust drone detection systems is further complicated by the limited availability of annotated infrared (IR) datasets. Recent advances in generative models and simulation tools enable the creation of synthetic IR imagery that approximates real thermal scenes, offering precise control over object geometry and scene composition while providing automatic ground-truth annotations. This work investigates a synthetic-first training strategy for LWIR drone detection. We present a synthetic data generation pipeline and evaluate three training protocols across three model architectures. We also introduce a dataset analysis framework that that allows us to relate dataset characteristics to downstream detection performance. Our results show that synthetic pre-training provides a useful addition to (few-shot) real-data fine-tuning but does not replace in-domain LWIR data. Models trained exclusively on synthetic data show limited generalization to real thermal imagery. In some RF-DETR configurations, incorporating RGB bird imagery during pre-training improves detection performance by increasing the geometric diversity of aerial objects. The strongest overall result in terms of mAP@50:95, is achieved by RF-DETR-L with dual-modality fine-tuning, although the difference from single-modality fine-tuning is small and should be interpreted cautiously. Even small amounts of real thermal imagery substantially improve detection accuracy by aligning learned representations with the statistical properties of physical sensors. Our experiments suggest that dataset alignment may have a strong impact on performance, although formal statistical correlation analysis is not performed. Our results suggest that semantic alignment in feature space is a useful diagnostic indicator of performance trends, while radiometric properties such as entropy and dynamic range provide additional context for detection robustness. This work provides a foundation for combining synthetic and real IR data for effective G2A drone detection.

## 1. INTRODUCTION

Ground-to-Air (G2A) drone detection in long-wave infrared (LWIR) spectral bands presents distinct challenges compared to object detection in the visible spectrum. Although infrared (IR) sensing enables operation in low-light conditions and degraded visibility environments, it introduces several modality-specific limitations, including reduced texture information, sensor noise, radiometric variability, and weak thermal contrast between airborne targets and complex backgrounds [1]. These factors make reliable detection of small aerial objects particularly difficult in operational scenarios.

The development of robust IR drone detection systems is further complicated by the limited availability of annotated LWIR datasets. Real-world IR data acquisition is costly and often restricted by operational constraints, while manual annotation of small airborne targets is time-consuming and prone to error [2]. As a result, conventional supervised learning approaches that rely on large-scale labelled datasets are difficult to apply in this domain. Synthetic data generation presents a promising alternative for scaling training data while reducing manual annotation costs.

Recent advances in generative models and simulation tools enable the creation of synthetic IR imagery that approximates real thermal scenes [3], [4], [5]. Synthetic datasets allow precise control over object geometry, pose variation, environmental conditions, and scene composition, while also providing automatically generated ground-truth annotations [6]. However, detectors trained solely on synthetic data often suffer from a domain gap when applied to real IR imagery [3], [6], [7]. Differences in radiometric characteristics, sensor noise patterns, and thermal diffusion effects can lead to discrepancies between simulated and physical data, which in turn limits model generalizability [7], [8], [9].

This study investigates a synthetic-first training strategy for LWIR drone detection that combines synthetic scene generation with targeted real-data fine-tuning. Synthetic IR scenes are generated through a structured simulation pipeline that models drone geometry and pose variation while approximating thermal-like target appearance and incorporating diverse environments as backgrounds. These scenes serve as the primary source of training data and enable scalable detector pre-training without manual annotation.

To evaluate the effectiveness of different (synthetic) training datasets, this work also introduces a dataset analysis framework designed to quantify the alignment between synthetic and real LWIR imagery. The framework evaluates datasets at multiple levels, including pixel-level radiometric characteristics, representation-space alignment within the detection model's feature space, and structural dataset diversity. By linking these properties to downstream detection performance, the analysis provides insight into how dataset composition influences model robustness.

*tanel.liiv@marduk.ee

The main contributions of this work are fourfold. First, we develop a structured synthetic data generation pipeline for G2A drone detection in IR imagery, combining IR-like background generation, Blender-based scene construction, and post-processing to simulate sensor noise, while providing automatic ground-truth annotations. Second, we evaluate a synthetic-first training strategy across multiple detector families, examining how synthetic pre-training combined with limited real LWIR fine-tuning affects detection performance under severe real-data scarcity. Third, we introduce a dataset analysis framework that characterizes synthetic and real thermal datasets through pixel-level radiometric statistics, semantic feature-space alignment, and structural diversity metrics. Fourth, we analyze the incorporation of bird imagery as a hard-negative in training data, providing evidence that additional aerial-object morphology can support drone detection in some detector configurations while also introducing potential domain mismatch.

## 2. RELATED WORK

Training G2A drone detection models in LWIR imagery with synthetic data presents distinct challenges. Solutions to these challenges can be found in the adjacent fields of IR and/or small object detection, thermal dataset curation, synthetic data generation, and cross-domain adaptation. In this section, we thus elaborate on the state-of-the-art of LWIR and small object and drone detection, training of detectors with synthetic data, generation of synthetic data, and evaluating the quality of synthetic data.

### 2.1 Object detection in LWIR

In recent years, major advances have taken place in the field of object detection, such as the emergence of CNN- (e.g. Faster R-CNN [10], YOLO [11]) and Transformer-based (e.g. RF-DETR [12]) architectures, enabling real-time detection for a wide range of tasks [11], [13]. Transformer architectures in particular have shown promising results in automatic target detection [14]. Compared with detection in the visible spectrum, LWIR detection is constrained by lower spatial resolution and signal-to-noise-ratio, atmospheric scattering, background noise, limited texture information and variability in sensor sensitivity [15], [16]. This makes performance depend not only on the detector ability to capture small objects in noisy backgrounds, but also on how closely the training data matches the physical and statistical properties of the deployment sensor. Overall, annotated LWIR drone data remains costly, sensor-dependent, and difficult to scale across ranges, backgrounds, drone types, and operating conditions.

### 2.2 Small target and drone detection

Low signal-to-noise ratio, clutter, weak target structure, and thermodynamically varying signatures are core challenges in small and dim IR target detection, especially when drones occupy only a few pixels and appear as low-texture silhouettes [1], [16]. Dataset studies conclude that for training object detectors data and annotation quality and curation matter more than dataset volume [2]. In these datasets, the inclusion of multi-sensor data, visible/IR coverage and the inclusion of hard negatives, e.g. birds, airplanes, and helicopters, contribute to improved final object detector performance [17], [18]. The inclusion of birds as hard negatives is a common practice for recent drone detection datasets [17], [19], [20], motivated by the resemblance between drones and birds in apparent size, silhouette, and trajectory. Recent work, including REDETR-RISTD [21] and RT-DETR adaptations for small objects [22], has also shown that architecture changes can improve detector performance on small targets. Finally, exploiting temporal information can also improve small object detection in a military context [23].

### 2.3 Synthetic data for training object detection models

Synthetic data offers a practical route for reducing dependence on scarce real datasets and annotations, offering perfect ground truth annotations and the ability to vary object placement, backgrounds, rendering parameters, distractors, post-processing, and camera settings with a high level of control [24], [25], [26], [27]. Recent work shows that synthetic-to-real detection performance depends strongly on training design, data augmentation [28], domain randomization [29] architecture choice, and dataset diversity [27] and mixing of real and synthetic data [30], with Transformer-based architectures showing promising results for training on synthetic data [28]. Remote-sensing generation methods such as AeroGen [31] add an important perspective: synthetic images must preserve not only image-level realism but also object scale, layout, localization, and annotation consistency. For LWIR drone detection, this is especially important because background diversity and thermal realism are separate requirements. A pipeline may generate varied scenes while still failing to reproduce realistic target heat distribution, sensor noise, and contrast behavior.

### 2.4 Strategies for synthetic data generation

Attempts at visible-to-IR translation and IR synthesis further show that the creation of synthetic thermal data should not be treated as ordinary style transfer. F-ViTA [3] and DiffV2IR [32] use semantic and wavelength-aware modelling to preserve scene structure during visible-to-thermal translation. Physics-aware work, including Mao et al. [33], argues more directly that IR image formation depends on material emissivity, heat distribution, optics, atmospheric attenuation, sensor response, non-uniformity correction, compression, and Automatic Gain Control (AGC). Javidnia [9] also shows that 16-bit-to-8-bit thermal tone mapping can materially affect object detection performance. These studies define the main limitation of the present synthetic pipeline: it produces scalable IR-like training data with controllable geometry, pose, background diversity, and automatic labels, but it is not a fully radiometric LWIR simulator. Real LWIR fine-tuning therefore remains necessary for sensor-domain alignment.

### 2.5 Evaluating the quality of synthetic data

The reality gap, the difference between real and simulated data, is often caused by a lack of diversity, poor domain fit, and low fidelity, and severely limits usability of synthetic data [27]. In LWIR imagery in particular, the reality gap occurs at the pixel and spatial-frequency level, due to rendering engines differing from real microbolometer sensors in modeling heat dissipation, dynamic range compression, and sensor noise [34]. Measuring the reality gap can be done using several metrics [35], some focusing on pixel-level variations, such as the Earth Mover's Distance (EMD), commonly used for cross-modality alignment [36], [37], [38], Gray-Level Co-occurrence Matrix (GLCM) statistics [39], capturing micro-textures of LWIR, Sobel gradients [40], capturing intensity fluctuations. Other metrics focus on capturing semantic differences, such as the Kernel Inception Distance (KID) [41] and Frechet Inception Distance (FID) [42].

## 3. METHODOLOGY

To achieve the research objective, the proposed methodology, illustrated in Figure 1, followed a sequential pipeline that included the generation of synthetic IR scenes, post-processing, and training of the state-of-the-art and legacy object detector models to fit the G2A drone detection task in the IR spectrum.

### 3.1 Synthetic IR image generation

In the military domain, publicly available, sufficiently large, sensor-diverse, and consistently annotated IR datasets for training supervised detection models remain limited. Therefore, a common approach in the literature, which we adopted, is to create a synthetic scene/image dataset that approximates military operational conditions and use it to train object detectors [43], [44]. Our synthetic IR scene generation process involved (1) generating high-resolution IR backgrounds and (2) creating low-resolution synthetic IR scenes. We lowered the resolution because IR cameras often have a significantly lower spatial resolution than visible-light cameras [45], [46].

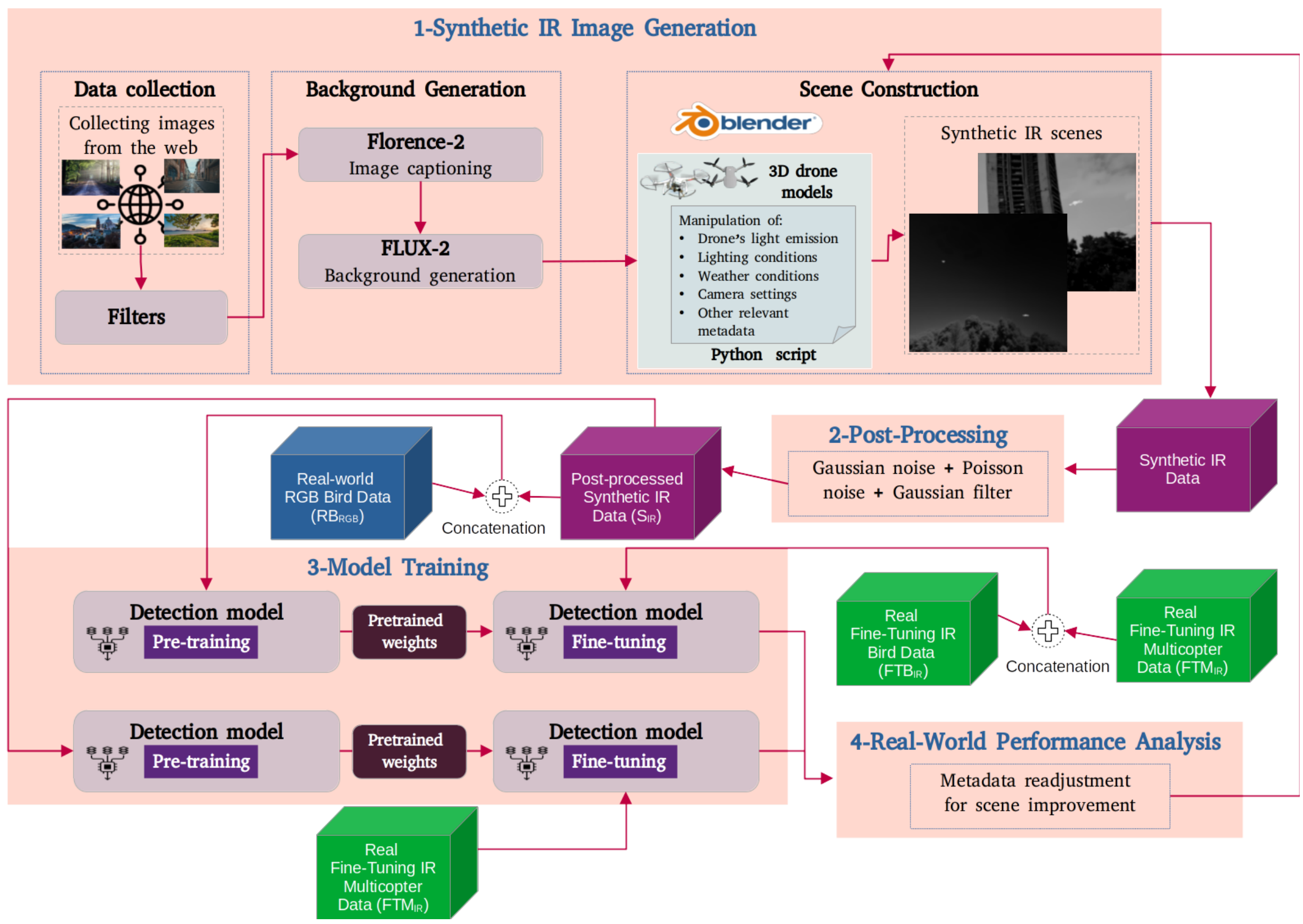

Figure 1: Synthetic IR generation pipeline for G2A drone detection training. In this context, IR specifically refers to long-wave infrared (LWIR).

**Background generation:** to generate the backgrounds, we initially collected 2,000 background images from the web. We employed scripts that automatically searched and downloaded photos from Pexels and Pixabay APIs. To generate a diverse set of search queries, we created pools of keywords for environments, weather conditions, and camera angles. The scripts combined keywords from each pool to search for a wide variety of scenes. Duplicate images were excluded and the following criteria were applied: (1) low human presence, reflecting the sparsity of civilians in a war zone; (2) suitable perspective, to simulate a large distance between the camera and the drone, ranging from 30 to 1,000 meters and the illusion of a 3D space; and (3) heterogeneous environment and weather conditions. After filtering, 1,751 images were retained and processed by two stacked generative models, Florence-2 [47] and FLUX [48], to generate the backgrounds. Florence generated detailed captions from the web images, and these captions were fed into FLUX to generate false-color IR backgrounds with a size of 3072 x 1536 pixels. FLUX was primarily designed for visible-spectrum image generation rather than physically accurate LWIR radiometric simulation. Therefore, we augmented Florence's detailed captions with a positive prompt containing IR artifact descriptions to generate IR-like false-color backgrounds. Although this approach generally produced visually plausible IR-like backgrounds, it should not be interpreted as physically accurate thermal rendering. We observed irregularities, such as thermal object distribution issues and visible-light-generated pixels in a subset of the generated outputs. These samples were excluded from the pipeline, resulting in a final dataset comprising 1,362 false-color IR-like backgrounds.

**Scene construction:** the scene construction was carried out in Blender [49], free and open-source 3D creation software. By leveraging Blender's Python module (bpy), we programmatically manipulated 3D objects, camera settings, and the 3D environment to build synthetic scenes from previously generated backgrounds. The iterative scene creation process typically involved randomly selecting a generated background and integrating it into the 3D environment; randomly positioning and rotating the drone in the camera's field of view; creating a motion animation; fog simulation, random adjustment of camera parameters, such as focal length and aperture f-stop, and random positioning of the light source. To approximate thermal-like target appearance, we first mapped the false-color IR-like background into the grayscale domain, then applied a light-emission component in Blender to the drone objects. The emission strength was controlled using a random value to introduce variability in the thermal-like target appearance. The emission intensity was increased when the background was darker to maintain sufficient contrast between the object and the environment in the simulated IR imagery. Note that, since the synthesized scenes were 512 x 512 pixels in size, only a cropped region of each background (3072 x 1536) was rendered at each iteration. This allowed us to match the resolution of real-world IR images, which is often low, and to create different scenes from a single background. In total, 20K samples of synthetic IR scenes were created. Each contained up to 3 drones, resulting in a comprehensive dataset comprising 29,945 quadcopter instances, including ground truth annotations. Scene metadata was also stored for reproducibility and analysis purposes. Two examples of synthetic IR scenes are provided in Figure 2.

**Post-processing**: FLUX struggles to reproduce realistic sensor noise, often resulting in scenes with overly smooth and linear signals. This can be problematic because it hinders the ability of the object detector to learn from noisy training samples. The robustness of a detection system is generally assessed by its capacity to handle various types of noise, including capture noise, sampling noise, and image encoding noise. Therefore, it is essential to include distorted samples in the training phase. To achieve this, we applied various noise-based degradations, including additive Gaussian noise and Poisson noise, to a random subset of the generated scenes. Gaussian noise approximates electronic and read-noise-like perturbations, while Poisson noise approximates signal-dependent photon-counting variation during exposure.

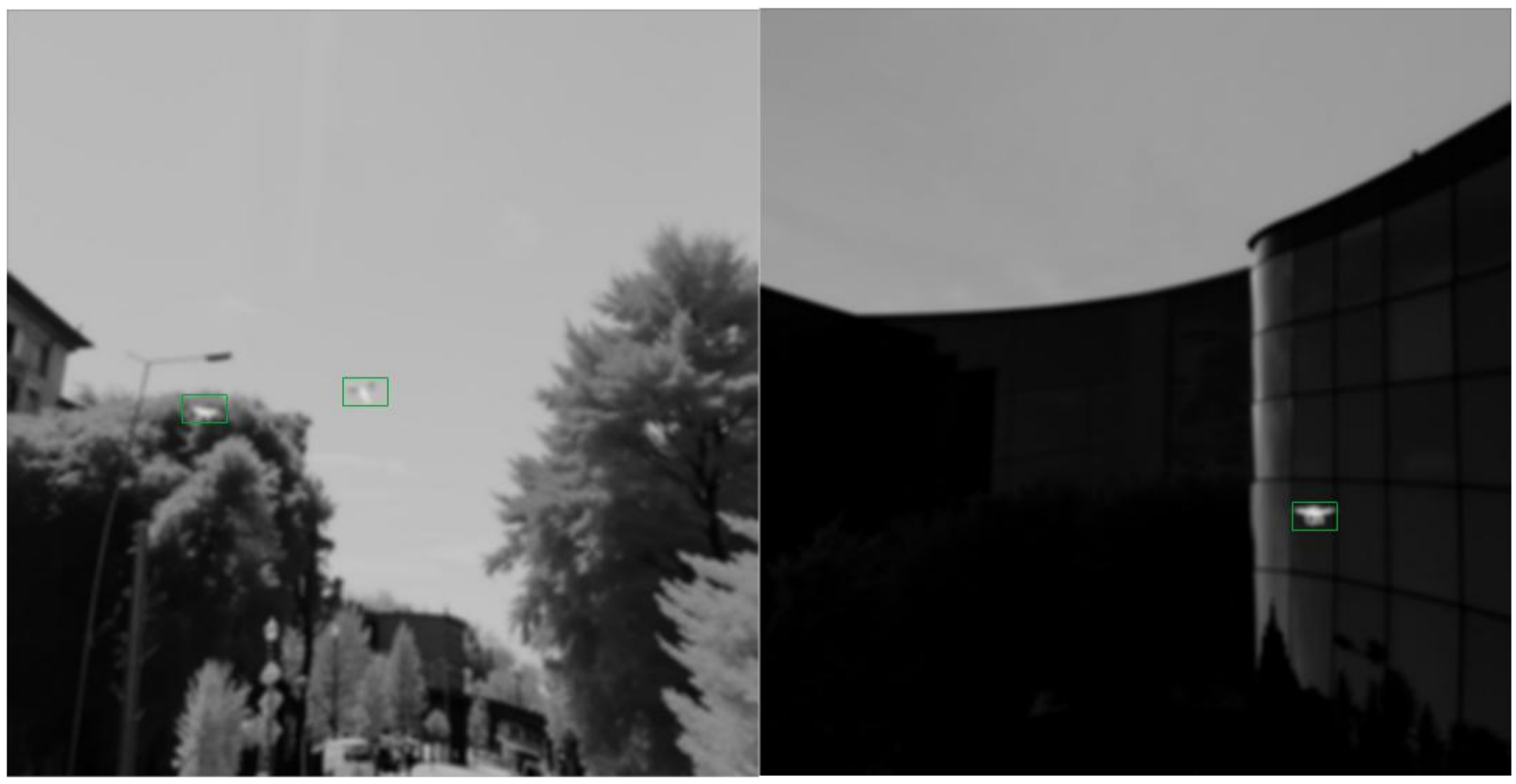

Figure 2: Examples of synthetic IR-like scenes. The bounding boxes are included for visualization purposes.

Additionally, Gaussian filtering with a 5 x 5 kernel was applied to simulate blurring effects that may arise from factors such as digital zoom. Additive Gaussian noise, Poisson noise, and Gaussian filtering were applied simultaneously or exclusively.

### 3.2 Models

Object detection frameworks are generally classified into three main paradigms: two-stage detectors, one-stage detectors, and transformer-based approaches. To ensure a representative evaluation across these paradigms, we employed two state-of-the-art detectors: the one-stage YOLO-family detector YOLOv13 [50] and the DETR-style transformer detector RF-DETR [12], alongside the widely adopted two-stage Faster R-CNN [10]. YOLOv13 is a relatively recent model and was selected over earlier YOLO variants due to its superior and more consistent performance across our preliminary experiments. The evaluation was conducted across multiple model scales, namely YOLOv13 nano, small, large, and extra-large, RF-DETR nano, small, medium, and large, and Faster R-CNN with a medium-sized ResNet backbone.

The training protocol included the following configurations:

1. Pre-training on synthetic IR ($S_{IR}$) data (20K samples) followed by fine-tuning on real-world IR multicopter[1] ($FTM_{IR}$) data (100 samples) which we called *single-modality fine-tuning*;
2. Pre-training on synthetic IR ($S_{IR}$) data (20K samples) + proprietary RGB bird ($RB_{RGB}$) data (19,875 samples), followed by fine-tuning on 100 real-world IR multicopter images ($FTM_{IR}$) and 100 real-world IR bird images ($FTB_{IR}$), for a total of 200 fine-tuning samples which we called *dual-modality fine-tuning*; and
3. Real-only training on real-world IR multicopter ($RM_{IR}$) data (100 samples), as a baseline.

The validation set included 3K samples of real-world IR multicopter data. The combination of IR and RGB images stems from the fact that some studies have shown that combining dual-modality spectral data can improve the performance of object detectors [51], [52]. Furthermore, we hypothesized that incorporating bird data would introduce challenging samples since bird morphology is often similar to drone geometry. This approach enhances the model's discriminative power and overall robustness. We evaluated the performance of the models by assessing mean Average Precision (mAP) across a range of IoU thresholds from 0.50 to 0.95. Additionally, Precision, Recall, and mAP@50 metrics are reported in the Appendix to provide further insight into model performance. For fixed-threshold Precision and Recall, the confidence threshold and IoU threshold were set to 0.001 and 0.7, respectively.

All models were initially pre-trained on the MS COCO dataset and that, consequently, our pre-training stage is actually a second round of pre-training. All models were trained on NVIDIA A100 GPUs with a batch size of 16. To ensure optimal performance without overfitting, training schedules and hyperparameters were tailored to the established best practices of each respective architecture. While the YOLOv13 models were trained for 100 epochs to accommodate their heavy augmentation pipelines, the two-stage CNN and transformer architectures require significantly shorter convergence windows. For the two-stage CNN comparison, we utilized the Detectron2 implementation of Faster R-CNN (ResNet-50 FPN, 3x schedule). We capped its training at 30 epochs, closely mirroring Detectron2's standard 3x schedule (~36 epochs), which is the established upper bound for full convergence under standard augmentation regimes. In parallel, the four RF-DETR variants were also trained for 30 epochs. Recent studies demonstrate that RF-DETR natively converges exceptionally fast, often plateauing within 10 epochs [53]. Consequently, these architecture-specific training regimes provide ample runway for full model convergence while maximizing computational efficiency. Because each detector was trained using its architecture-native recipe, random seed control and checkpoint-selection procedures were not fully harmonized across architectures. Therefore, comparisons between different detector families should be interpreted as indicative rather than as strictly controlled architecture ablations. The main conclusions are drawn primarily from within-architecture comparisons across training-data configurations. The complete matrix of training parameters is detailed in Appendix A.

## 4. DATA

This section details the datasets used in our experiments and establishes a statistical framework to evaluate them. We first outline the composition of the real and synthetic thermal datasets. To provide deeper understanding of the downstream detection performance, we then introduce hierarchical metrics spanning from pixel-level radiometry to high-dimensional feature space alignment. By quantifying both the structural diversity of these datasets and their visual semantic domain gap relative to the real-world validation baseline we establish a comprehensive understanding that contextualizes the subsequent model benchmarking results.

### 4.1 Datasets

We curated several datasets for this study, ranging from smaller real-world sets to high-volume synthetic sets. These are summarized in Table 1. The validation and fine-tuning sets were constructed from six public real-world LWIR datasets

[1] Quad- and hexacopters were included in this dataset.

[17], [18], [54], [55], [56], [57]. All real data utilized in this study was provided as processed 8-bit images, not raw radiometric data, aligning with the standard input of object detectors. To introduce distractors, we incorporated two distinct bird datasets: a proprietary visible-spectrum (RGB) dataset paired with the synthetic data for pre-training, and a curated set of real IR bird images from one of the public datasets utilized during fine-tuning. Birds were not included in the validation set; their sole purpose was to serve as hard negatives for model training. Full dataset compositions, including the drone model breakdown and frame-sampling strategies, are catalogued in Appendix B. We restricted our fine-tuning benchmarking to the 100-sample subset to reflect the data fragmentation and scarcity caused by differences in thermal sensors and drone models. This tests whether a limited data budget can successfully bridge the reality gap. Our subsequent domain analysis confirmed this subset closely matches key domain-alignment metrics of the broader dataset, particularly cropped object feature-space distance, but it may still inherently underrepresent object-level diversity.

Table 1. Overview of the datasets used in this study.

| Dataset Category | Total Images | Multicopter Images | Multicopter Instances | Bird Images | Bird Instances |
|---|---|---|---|---|---|
| Validation Set [18], [54], [55], [56] | 3,000 | 3,000 | 3,000 | 0 | 0 |
| Real Fine-Tuning Multicopter ($FTM_{IR}$), source set [17], [56], [57] | 2,700 | 2,700 | 2,871 | 0 | 0 |
| Synthetic ($S_{IR}$) (own contribution) | 20,000 | 20,000 | 29,945 | 0 | 0 |
| Synthetic + RGB Birds ($S_{IR}$ + $RB_{RGB}$) | 39,875 | 20,000 | 29,945 | 19,875 (RGB) | 35,000 (RGB) |
| Real Fine-Tuning Multicopters, selection from [17], [56], [57] ($FTM_{IR}$) | 100 | 100 | 104 | 0 | 0 |
| Real Fine-Tuning Multicopters, selection from [17], [56], [57] ($FTM_{IR}$) + IR Birds ($FTM_{IR}$ + $FTB_{IR}$) | 200 | 100 | 104 | 100 (IR) | 145 (IR) |

### 4.2 Dataset Analysis Methodology

SDQM [58] argues that generated datasets should be assessed through diagnostics related to downstream detection utility. Following this principle, we analyzed datasets at three levels: (1) pixel-level radiometric statistics, (2) semantic feature-space alignment using Kernel DINO Distance, and (3) structural diversity using the Vendi Score. These metrics are used as diagnostic tools rather than as formal causal predictors, helping to interpret whether performance differences arise from radiometric mismatch, feature-space distance, insufficient diversity, or the interaction between these factors. The analysis was performed on both full images and isolated target crops.

At the pixel level, we established metrics to capture the fundamental radiometry and structural properties of the images. For example, we utilize Histogram Total Variation (HTV) to measure the fragmentation of pixel brightness values in the images by calculating the cumulative difference between adjacent histogram bins (visualized in Figure 3, top row). To evaluate the target-crop scale, we employ Sobel Gradient Variance to measure the intensity of sharp thermal transitions of the target objects (Figure 3, bottom row). Full details and other pixel-level metrics not discussed in the main text are provided in Appendix C.

To measure the visual semantic domain gap, we introduce the Kernel DINO Distance (KDD)—an adaptation of the Kernel Inception Distance (KID) [41], calculated over the feature space of the DINOv2 [59] vision transformer. KDD provides a feature-space proxy for the statistical distance between two datasets, estimating how far a dataset diverges from the real-world validation baseline. To evaluate diversity within this same feature space, we utilized the Vendi Score [60], which estimates the effective number of unique visual concepts within a single dataset, providing a robust measure of internal variety. The complete implementations and theoretical justifications for these metrics, including our methodology for mitigating sample size bias across vastly different dataset volumes, are provided in Appendix C.

### 4.3 Pixel-level physics and hardware divergence

At the full-image level, real LWIR imagery exhibited higher Histogram Total Variation (HTV). In fact, this HTV disparity represents the largest measured effect size in our analysis between the real validation set and synthetic ($S_{IR}$) data (Cohen's $d = 1.242$). This elevated roughness (Figure 3, top left) was consistent with AGC-like stretching, which remaps high-bit-depth thermal measurements into an 8-bit display range, while maximizing dynamic range. The synthetic pipeline did not yet reproduce this scene-adaptive hardware behavior. However, our application of post-processing noise to synthetic data effectively approximated the noise texture profile of real validation data, yielding highly comparable median full-image texture metrics (full details in Appendix D).

At the target-crop level, real drones displayed significant structural complexity due to uneven heat distribution, sensor noise, and optics—factors largely absent in the comparatively uniform meshes of the synthetic pipeline (Figure 3, bottom row). This difference is shown by the Sobel Gradient Variance, which is over four times higher in real detections than in synthetic ones ($d = 1.224$). Furthermore, as detailed in Appendix D, substantial differences exist even among the real datasets, demonstrating that sensor hardware and AGC logic drastically alter how targets look, worsening the thermal data scarcity problem because target appearance becomes sensor-dependent.

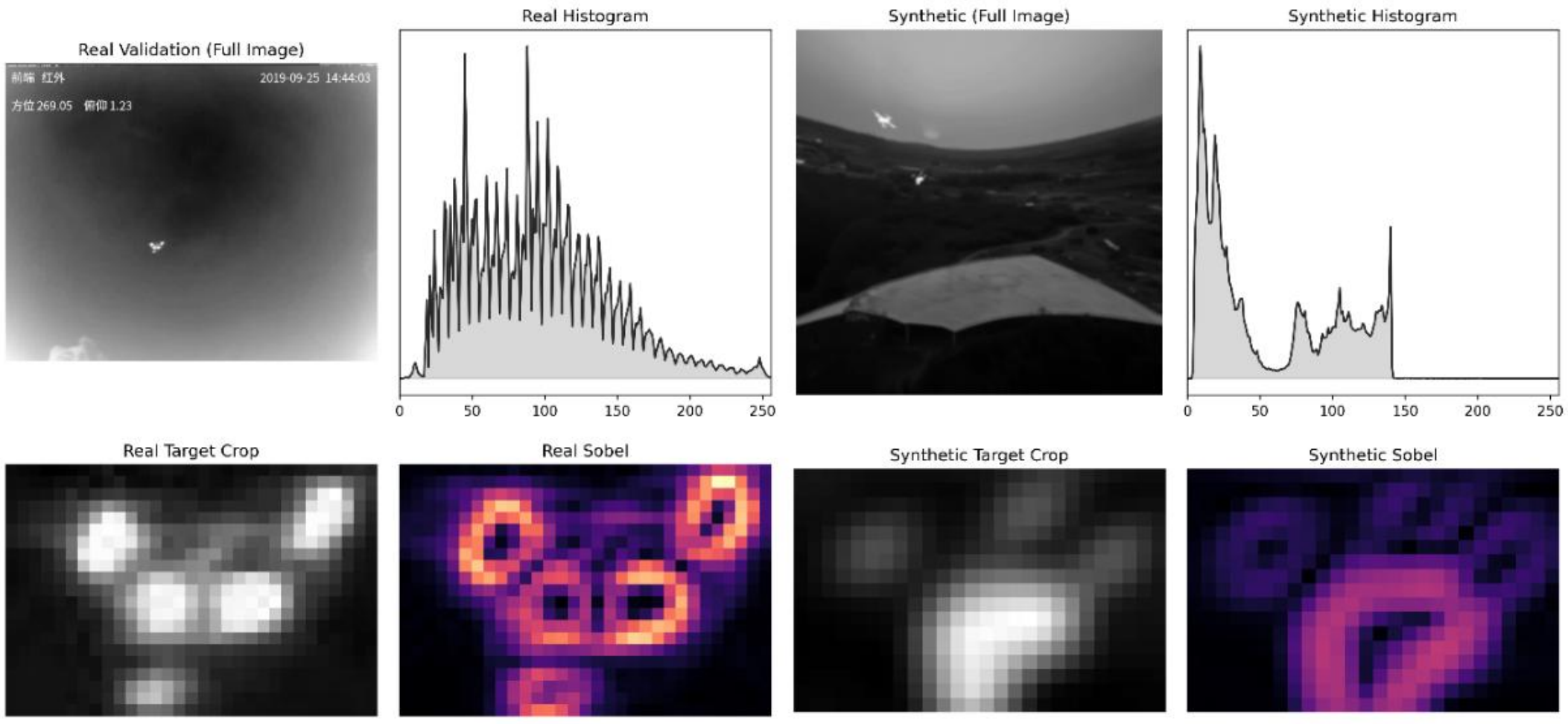

Figure 3: Comparison of pixel-level radiometry and structural complexity between a real validation and synthetic ($S_{IR}$) sample. Top: Full-image comparison; real imagery exhibits high Histogram Total Variation (HTV) due to AGC contrast stretching, whereas synthetic data is smoother. Bottom: Target-crop comparison showing Sobel gradient magnitudes; real drones display high Sobel Gradient Variance due to uneven heating and sharp thermal transitions, contrasting with more uniform synthetic meshes.

### 4.4 Visual Semantic Domain Gap and Structural Diversity

Beyond raw pixel statistics, the most notable divergence arises in visual semantic feature space. In the physical world, thermal backgrounds naturally wash out into low-contrast, low-variance states as scenes approach thermal equilibrium. Because our synthetic pipeline generates highly varied environmental scenes, the resulting imagery retains dense, high-frequency geometric details, such as highly textured terrain and foliage. This artificially inflates the full-image structural diversity (Vendi Score) far beyond what a physical thermal sensor typically captures, driving a substantial semantic domain gap (KDD) relative to the real validation set.

To evaluate drones separately from these environments, we also analyzed the target crops. At the target-crop scale, the Vendi Score of the synthetic targets drops significantly compared to the full images, aligning closely with the real-world validation targets. This confirms that the inflated full-image diversity in the synthetic data is driven almost entirely by the backgrounds rather than the drones. However, while the synthetic pipeline achieves realistic structural diversity at the object level, its feature-space distance (KDD) remains substantially higher than the real data. This indicates that matching morphological diversity alone is insufficient to close the thermal domain gap. This dichotomy is effectively visualized in Figure 4, which plots both the full-images and the target crops within the KDD-Vendi feature space. It also validates the 100-sample fine-tuning sampling strategy, confirming it roughly corresponds to the full set's characteristics. The complete KDD/Vendi quantitative breakdown is detailed in Appendix E, while Appendix F confirms that the observed dataset diversity hierarchy is immune to sample size.

The inclusion of birds introduces a distinct shift in feature-space alignment. As shown in Figure 4, appending the large-scale RGB bird dataset (orange square) to the pure synthetic baseline results in a measurable increase in object-level KDD. Conversely, the inclusion of IR birds (pink diamond) induces a much smaller feature-space shift relative to its base fine-tuning set, maintaining a closer alignment to the validation baseline. This suggests that while visible-spectrum imagery provides helpful morphological variety, it also introduces cross-modal divergence that moves the distribution further from the LWIR target domain. Importantly, because these two bird sets are integrated into different base data configurations and later applied at different stages of the training pipeline (pre-training vs fine-tuning), this does not represent an isolated RGB-vs-IR ablation study. Rather, it illustrates how each specific distractor modality independently shifts the feature-space distribution of its respective base dataset.

The overall trajectory is clear: while the generator produces highly accurate foreground morphologies, the combination of highly complex environments and cross-modal distractors shifts the resulting distribution significantly away from the real-world baseline. To understand how these structural and radiometric domain gaps translate into actual operational performance, Section 5 benchmarks the accuracy of downstream detection models trained across these configurations.

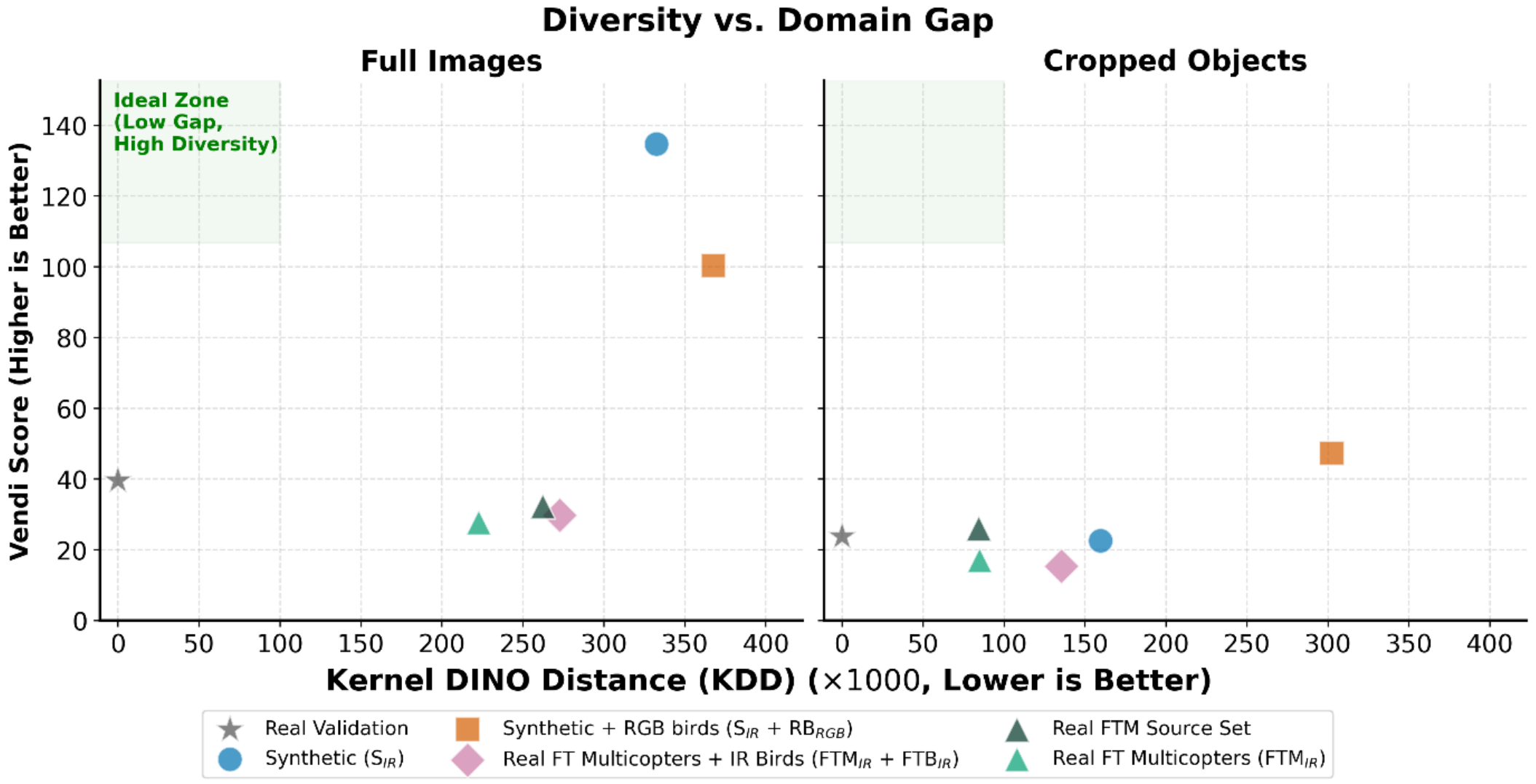

Figure 4. Diversity and domain gap visualization. The shaded green region denotes a hypothesized desirable region rather than a guaranteed optimum.

## 5. RESULTS

This section presents the experimental results of the proposed IR drone detection pipeline and evaluates the impact of the single-modality and dual-modality fine-tuning strategies described in Section 3 using the real-world LWIR validation dataset introduced in Section 4.1. Figure 5 shows the mAP@50:95 values of the models according to the different training strategies. Absolute results across all metrics including hardware performance and inference latency are reported in Table 8 (Appendix G) and Figure 10 (Appendix H). The key findings of our experiments are as follows.

**RF-DETR-L with multi-modality fine-tuning reaches the best performance:** the highest overall mAP@50:95 was achieved by RF-DETR-L with multi-modality fine-tuning, reaching 0.723, closely followed by RF-DETR-L with single-modality fine-tuning at 0.713. Since this difference is small and repeated-seed experiments were not performed, it should be interpreted cautiously. Faster R-CNN showed the largest relative gain because its real-only baseline was weak, while YOLOv13 also benefited from synthetic-first training.

**Synthetic ($S_{IR}$) + real-world RGB bird ($RB_{RGB}$) pre-training outperforms synthetic-only pre-training:** across all model architectures, the introduction of the RGB bird dataset (as hard negatives) led to an improvement of the mAP@50:95 compared to synthetic-only pre-training. The KDD distance in Figure 4 demonstrates that combining S and $RB_{RGB}$ data increases the domain gap compared to real-world IR multicopter data. This suggested that adding RGB bird imagery moves the dataset further from the LWIR validation distribution in the chosen feature space, likely due to the deviating modalities. Therefore, we assume that the performance improvement stems primarily from the morphological similarity between drones and birds, which strengthens the model's discriminatory power and improves its robustness.

**Real LWIR fine-tuning is necessary for sensor-domain alignment:** with the exception of Faster R-CNN, for which we hypothesize the architecture struggles to effectively handle low-resolution thermal features, none of the synthetic pre-training strategies achieve higher performance than the real-only baseline. This suggests that the diversity of latent features introduced mainly by backgrounds (see Vendi score in Figure 4) in the $S_{IR}$ and $S_{IR}$+$RB_{RGB}$ datasets does not provide accurate thermal signals.

**Synthetic pre-training improves few-shot LWIR drone detection:** Figure 5 shows that synthetic pre-training followed by real-data fine-tuning improves few-shot (100 samples) LWIR drone detection performance across the evaluated detector families compared to the real-only baseline. An improvement in mAP@50:95 is observed across all models under the single-modality fine-tuning configuration, demonstrating that the generated IR-like synthetic scenes provide useful pre-training signal for the G2A drone detection task. Faster R-CNN exhibits the lowest mAP@50:95 when trained solely on real-world data (0.012) and both the single-modality and dual-modality fine-tuning approaches achieve the largest gains for this architecture, at 0.352 and 0.278, respectively.

**Bird distractors have a conditional fine-tuning effect**: RF-DETR variants benefited consistently though slightly from the addition of bird distractors to the training data. YOLOv13-s also showed slight performance gains, while YOLOv13-n, YOLOv13-l, YOLOv13-xl and Faster R-CNN did not seem to benefit. Although birds are relevant because they share aerial-object morphology with drones and can function as hard negative samples, they introduce cross-modal mismatch relative to the LWIR target domain. Consequently, certain architecture–scale combinations may struggle to overwrite or suppress irrelevant RGB bird features during fine-tuning from a relatively small thermal image dataset. This can limit their ability to learn modality-specific representations, which are more relevant for LWIR drone detection.

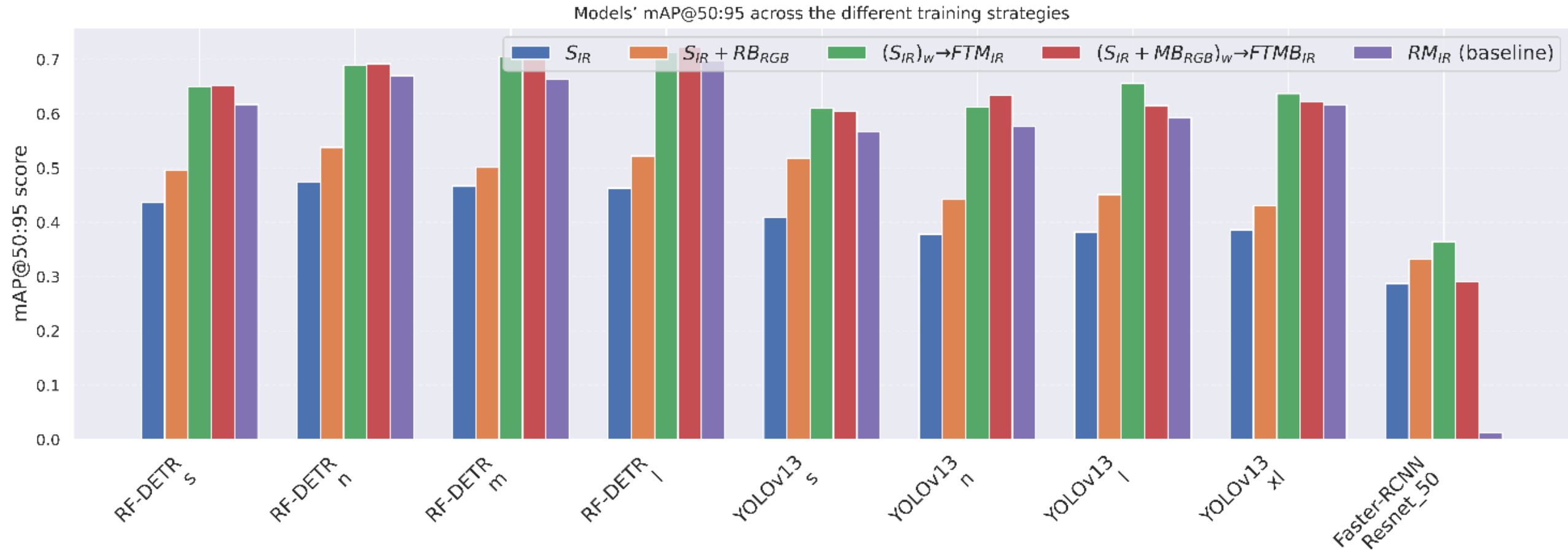

Figure 5: Model performance. $S_{IR}$: pre-training on Synthetic data; $S_{IR} + RB_{RGB}$: pre-training on Synthetic and proprietary RGB bird data; $(S_{IR})_w \rightarrow FTM_{IR}$ : weights were initialized from pre-training on Synthetic data, then fine-tuned on real-world LWIR multicopter data; $(S_{IR} + RB_{RGB})_w \rightarrow FTMB_{IR}$: weights were initialized from pre-training on Synthetic and proprietary RGB bird data, then fine-tuned on real-world LWIR multicopter and bird data; $(S_{IR})_w \rightarrow FTM_{IR}$ : weights were initialized from pre-training on Synthetic data, then fine-tuned on real-world LWIR multicopter data; R: the models were trained solely on real-world IR multicopter data.

## 6. DISCUSSION

This study evaluated a synthetic-first training strategy for ground-to-air drone detection in LWIR imagery under severe real-data scarcity. The main finding of this work is that synthetic IR-like pre-training provides a useful addition to (few-shot) real-data fine-tuning but does not replace in-domain LWIR data. Across the evaluated detector families, synthetic pre-training followed by fine-tuning on a small real LWIR dataset improved mAP@50:95 relative to training only on the limited real set, with RF-DETR-L reaching the highest overall performance in this set-up. This indicates that an effective fine-tuning approach can substantially reduce the limitations associated with limited real-data training, even for a two-stage detector architecture. Additionally, synthetic-first training with limited real LWIR fine-tuning can outperform real-only learning in a data-constrained G2A drone-detection benchmark. However, although synthetic data are beneficial for improving model initialization and generalization, they are not a substitute for real-world LWIR data.

Our dataset analysis explains why synthetic data are useful but insufficient alone and represents a second contribution of this work. The synthetic pipeline captures object-level morphology but does not fully reproduce the physical and sensor-dependent properties of real LWIR imagery, including target-level heat structure, sensor noise, optics, compression, and camera-side processing. The KDD, Vendi and radiometric analyses therefore indicate that morphological diversity alone is not enough to close the thermal reality gap. Even a small amount of real LWIR data is valuable and necessary because it aligns the learned representations with sensor-specific image statistics.

Bird imagery should be treated as a conditional robustness mechanism rather than a universally beneficial augmentation source. Birds provide relevant aerial-object morphology and can function as hard negatives, and several RF-DETR variants benefited from the bird-augmented configuration. However, the effect was not consistent across all detector families or scales, and RGB bird imagery also introduced cross-modal divergence from the LWIR target domain. In conclusion, bird data can serve as valuable negative samples; however, the modality gap between RGB and LWIR imagery may have a beneficial or detrimental effect, depending on the ability of a given architecture-scale combination to adapt learned representation to the thermal domain.

RF-DETR achieved the strongest absolute results in this setup. One possible explanation is that transformer-based detectors may benefit from long-range feature interactions and broader target–background context, which prior work has shown to be important in infrared small-target detection [1]. However, because the detector families were trained with architecture-native and non-harmonized recipes, this remains a plausible interpretation rather than a definitive conclusion.

### 6.1 Limitations and future work

This study has four main limitations, which we present alongside suggestions for future work.

First, the synthetic pipeline produces IR-like imagery rather than physically accurate LWIR simulation. Future work should incorporate physically grounded thermal rendering, target-level heat-distribution modeling, and sensor-specific AGC simulation.

Second, our validation setting is limited in sensor diversity, environmental conditions, target ranges, drone types, bird distributions, and deployment scenarios. Thus, the results should not be interpreted as demonstrating operational robustness in deployment environments. Further evaluation with independent sensors, broader environmental conditions, and additional real-world LWIR and MWIR data is required to establish reliable operational performance. Additionally, studies of operational robustness might consider real-time processing constraints, which are critical for functional drone detection systems. We incorporate performance and inference latency benchmarks for our models in Appendix H but future work might extend on this aspect of model evaluation.

Third, the architecture comparison was not a fully harmonized detector ablation, and repeated-seed evaluation was not performed. Future work should include controlled within-architecture ablations, repeated runs, learning curves across real-data budgets, and radiometrically better aligned hard-negative data.

Fourth, the limited number of dataset configurations prevented a robust correlation analysis between dataset analysis metrics and model accuracy, restricting our analysis to relative performance trajectories. Future work should evaluate a wider selection of pre-training and fine-tuning configurations to establish statistically robust correlations. Despite these limitations, the consistent within-architecture gains support the main conclusion that synthetic-first training is beneficial under the evaluated data-scarce LWIR benchmark conditions.

## ACKNOWLEDGEMENTS

This work received funding from the European Defence Fund through the project STORE (Shared daTabase for Optronics image Recognition and Evaluation), grant agreement №101121405. We would like to particularly express our gratitude to all consortium partners involved in the data acquisition and processing.

## APPENDIX A: MODEL TRAINING AND HYPERPARAMETER CONFIGURATIONS

To ensure the reproducibility of our experiments and to facilitate future comparative studies, the complete matrix of training hyperparameters is detailed in Table 2.

Table 2. Hyperparameter configurations for benchmarking models.

| Parameter | YOLOv13 | Faster R-CNN (ResNet-50) | RF-DETR |
|---|---|---|---|
| **Input Size** | 640 x 640 | Default: Dynamic Resize (Shortest Edge: 640–800px, Max: 1333px) | Scale-dependent (Defaults):<br>• Nano: 384 x 384<br>• Small: 512 x 512<br>• Medium: 576 x 576<br>• Large: 704 x 704 |
| **Optimizer** | Adam (Weight Decay: 0.0005) | SGD (Detectron2 Default) | AdamW (Architecture Default) |
| **Base Learning Rate** | 0.001 | 0.001 (with default linear warmup) | 0.0001 (Architecture Default) |
| **Scheduler** | Linear | Multi-step decay (66%, 85%) | Multi-step / Cosine (Architecture Default) |
| **Augmentations** | Mosaic (1.0), RandAugment, Erasing (0.4), Copy-Paste (0.1), FlipLR (0.5), Scale (0.5), Translate (0.1), HSV Jitter (Architecture Defaults) | Random Resize, Random Flip (Detectron2 Defaults) | Multi-scale Resize, Random Flip, Color Jitter (Albumentations Defaults) |
| **Training Duration** | 100 Epochs | 30 Epochs (Dynamically scaled to iterations) | 30 Epochs |
| **Batch Size** | 16 | 16 | 16 |
| **Pretrained Weights** | COCO | COCO | COCO |
| **Random Seed** | 0 (Fixed, Architecture Default) | Unspecified (Random) | 77 (Fixed) |
| **Confidence Threshold** | 0.001 (Architecture Default) | 0.05 (Detectron2 Default) | 0.001 (Architecture Default) |
| **NMS IoU Threshold** | 0.7 (Architecture Default) | 0.5 (Detectron2 Default) | N/A (NMS-free architecture) |
| **Evaluation Protocol** | Epoch-level Val Eval (Best Checkpoint saved via YOLO Fitness Score) | Terminal Eval (Final Checkpoint) | Epoch-level Val Eval (Best Regular Checkpoint saved via Peak mAP50:95) |

## APPENDIX B: DATASET CONFIGURATIONS

To complement synthetic data and enable real-world evaluation, we curated fine-tuning and validation sets from several public Long-Wave Infrared (LWIR) datasets. The real-world validation set was constructed using sequences from [18], [54], [55], [56]. Conversely, the real-world fine-tuning distributions were aggregated from [17], [56], [57]. All publicly available LWIR datasets utilized in this study were provided as processed 8-bit format images rather than raw radiometric data.

Because these datasets primarily consist of continuous image sequences captured during drone flights, frames were initially sampled at temporally spaced intervals to minimize redundancy. Severely degraded or low-quality frames were discarded. Following this extraction, the real-world frames underwent a rigorous manual curation and verification process. First, to prevent spatial or environmental bias during evaluation, the validation frames were filtered to ensure a balanced distribution of target positions and diverse background contexts. Second, to establish a reliable ground truth, all existing bounding box annotations across the real-world datasets were systematically reviewed and corrected.

To prevent data leakage, the fine-tuning and validation sets are constructed from entirely distinct datasets, with only two explicitly controlled exceptions. Although data from the Anti-UAV corpus and the Tavaris et al. [56] dataset were utilized in both phases, we enforced explicit separation criteria. For the Anti-UAV corpus, the fine-tuning data [57] and validation data [18] are drawn from two distinct capture campaigns. For the Tavaris et al. [56] dataset, although the data originates from the same broader collection effort, the validation subset is restricted to a different location and features a specific drone model that is entirely absent from the fine-tuning sequences drawn from this same dataset. While the data sources themselves are strictly isolated, the validation and fine-tuning sets intentionally include some overlapping drone models in order to evaluate target recognition across unseen sequences and backgrounds.

A 100-sample fine-tuning subset was extracted from the full real-world fine-tuning set using K-means sampling over a DINOv2 feature space. This subset was deliberately constructed to reflect the data fragmentation and scarcity inherent to real-world domain adaptation across varying thermal cameras and drone models. As validated by our later analysis, this K-means extraction closely preserves the structural diversity and domain characteristics of the overarching dataset, ensuring the severely constrained subset remains a mathematically robust baseline for evaluation.

The datasets were structured into distinct pre-training and fine-tuning phases to systematically isolate the effects of synthetic imagery, real-world thermal data, and visible-spectrum distractors. The complete quantitative composition of these splits, including modalities and sampling rules, is detailed in Table 3.

Table 3. Dataset configurations and compositions.

| **Role** | **Dataset Configuration** | **Total Images** | **Drone Instances** | **Bird Instances** | **Model Breakdown & Details** |
|---|---|---|---|---|---|
| Evaluation | Real-World Validation Set | 3,000 | 3,000 | 0 | DJI Phantom (mix of 2 & 4): 1,000 instances<br>DJI Inspire 2: 1,000<br>DJI Mavic Pro: 1,000 |
| Real Fine-Tuning (FT) Data | Real FT Multicopters ($FTM_{IR}$) Source Set | 2,700 | 2,871 | 0 | DJI Matrice 30, Aurelia X6, YUNEEC H520E, DJI Flame Wheel F450, DJI Phantom 4, DJI Inspire 2, DJI Mavic Pro |
| Only Multicopter Progression | | | | | |
| Pre-Training | Synthetic ($S_{IR}$) | 20,000 | 29,945 | 0 | DJI Inspire 2: 9,862<br>DJI Mavic Pro 3: 10,139<br>DJI Phantom 2: 9,944 |
| Fine-tuning | Real FT Multicopters ($FTM_{IR}$) | 100 | 104 | 0 | K-means sampled (via DINOv2 feature space) set from the Real FT multicopter source set. |
| Birds Progression | | | | | |
| Pre-Training | Synthetic + RGB Birds ($S_{IR}$ + $RB_{RGB}$) | 39,875 | 29,945 | 35,000 | Combines Synthetic ($S_{IR}$) with a proprietary visible-spectrum (RGB) bird dataset. |
| Fine-tuning | Real FTM + IR Birds ($FTM_{IR}$ + $FTB_{IR}$) | 200 | 104 | 145 | Combines $FTM_{IR}$ with matched-ratio real IR birds (also k-means sampled via DINOv2). |

## APPENDIX C: DATASET ANALYSIS METHODOLOGY

This analysis quantifies dataset alignment at three levels: pixel-level radiometry, representation-space semantics, and structural diversity. Because the dataset configurations differ substantially in size, direct comparisons may reflect dataset volume rather than domain characteristics. To mitigate this, certain statistics, such as luminance histograms, were normalized by pixel counts prior to aggregation, and distributional metrics were computed using sampling procedures designed to reduce sample-size bias. However, not all measurements can be fully normalized in this manner, and configurations containing the high-resolution RGB bird dataset may therefore exhibit distributional shifts driven partly by this native resolution disparity.

### C.1 Level I: Pixel-Level Radiometric and Textural Alignment

The primary domain gap between synthetic and physical LWIR imagery occurs at the pixel and spatial-frequency level. Rendering engines differ from real microbolometer sensors in modeling heat dissipation, dynamic range compression, and sensor noise [34]. To quantify this gap, all images were converted to single-channel 8-bit grayscale. Local Weber Contrast and Sobel gradient metrics were computed only for target bounding boxes, while other radiometric statistics were calculated over both full frames and isolated target crops. Crops smaller than 400 total pixels were excluded from distributional metrics to prevent bin-starvation bias. The 400-pixel threshold (20x20 spatial equivalent) was selected as a statistical floor because it provides approximately 1.5 times the number of pixels as there are bins in an 8-bit histogram (256). This ensures sufficient data mass to capture meaningful distributional shape and texture, whereas sub-threshold crops enforce mathematical sparsity and artificial quantization spikes.

### C.1.1 Radiometric Intensity, Quantization, and Contrast Alignment

Mean Luminance Distribution: To characterize brightness differences, we analyze mean luminance distributions using normalized histograms. Dataset-level histograms were averaged and compared using the Earth Mover's Distance (EMD), a Wasserstein metric commonly used for cross-modality alignment [36], [37], [38]. Because image luminance histograms are 1D arrays, we compute the exact closed-form 1D EMD (the $L_1$ distance between their Cumulative Distribution Functions).

**Quantization**: LWIR cameras typically compress 14-bit data to 8-bit via AGC, producing "comb-like" histogram artifacts. While simple sparsity metrics like Percentage of Empty Bins (PEB) can be misled by the "dithering" effect of microbolometer noise or the aggregation of multiple different sensors, we utilize Histogram Total Variation (HTV), a metric derived from the Total Variation principle [61]. HTV quantifies the cumulative absolute difference between adjacent intensity bins, providing a robust measure of histogram "roughness." High HTV values indicate the sharp, unnatural intensity jumps characteristic of aggressive AGC-driven quantization.

**Background vs Target Contrast**: Local target visibility was measured using Local Weber Contrast ($C_w$) [62]. The background region was defined as a hollow rectangle surrounding each bounding box with adaptive padding to maintain a 1:1 area ratio between target and background, ensuring scale-invariant contrast measurements.

### C.1.2 Signal-to-Noise and Texture Analysis

Real LWIR imagery contains micro-textures such as fixed-pattern noise and microbolometer grain. To quantify these characteristics, we computed Gray-Level Co-occurrence Matrix (GLCM) statistics [39], including Homogeneity, Correlation, and Entropy, averaged across four orientations to ensure rotation invariance.

### C.1.3 Intensity Gradient Distribution

Thermal image formation in real-world LWIR sensors is subject to germanium lens blur and thermal diffusion, which can naturally soften object boundaries. Conversely, internal physical heat sources and microbolometer noise may introduce localized intensity fluctuations that are often absent in the relatively uniform meshes of synthetic rendering. To investigate these competing spatial-frequency effects, we computed the variance of Sobel gradient magnitudes within target bounding boxes [40]. A noise-floor mask excluded near-zero gradients to avoid bias from empty regions.

### C.1.4 Dynamic Range Utilization

Thermal cameras dynamically stretch scene intensities using AGC. To evaluate how well synthetic imagery matches this behavior, we computed the effective dynamic range, defined as the difference between the 1st and 99th intensity percentiles. This excludes dead pixels and extreme noise spikes while capturing usable sensor bandwidth [34].

### C.2 Transitioning to Semantic Evaluation: Feature Extraction

While Level I evaluates raw pixel properties, semantic analysis requires embedding images in a neural feature space. Because our evaluation targets multiple distinct detection architectures, we utilized DINOv2 ViT-L/14-reg [59], [63] to establish a universal, model-agnostic baseline. Because targets typically occupy a small fraction of the frame, full-image embeddings largely reflect background environments. To isolate the object-level domain gap, embeddings were extracted from both full frames and isolated target crops. These multiscale embeddings serve as the foundation for our Level II and III metrics: Kernel DINO Distance (KDD) to quantify domain alignment, and Vendi Score to measure internal diversity (formally detailed in Sections C.3 and C.4).

To satisfy the model's 14×14 patch requirement, full images were processed at 644×644 because it is the closest multiple of 14 to the 640-pixel maximum dimension shared by the majority of our real LWIR datasets. Object crops were extracted with 10% contextual padding, forced to a square aspect ratio, and processed at 56×56. Because the primary structural divergence metric is measured relative to the real-world validation baseline, the extraction resolution must be anchored to preserve the structural integrity of this specific ground truth. The 56×56 resolution is the closest architectural patch-multiple to the validation set's native median target size (43.3 px, measured by the longest edge of the bounding box). Utilizing standard larger resolutions (e.g., 224×224) would require extreme upscaling for the synthetic targets and the infrared bird distractors (which feature median longest-edge sizes of 20.3 px and 9.8 px, respectively).

This heavy interpolation induces high-frequency artifacts that artificially inflate structural divergence. As demonstrated in Figure 6, while the relative hierarchy of the datasets remains largely stable across resolutions, increasing the crop size beyond 56 pixels causes a rapid inflation in both KDD and Vendi scores. This effect becomes particularly unstable at the larger 224×224 resolution, where the pure synthetic set (blue line) experiences a disproportionate surge. Because these synthetic targets are natively small, forcefully stretching them by an order of magnitude forces the model to evaluate severe upscaling noise rather than true morphological features. Ultimately, anchoring the evaluation at 56×56 ensures that the resulting metrics measure the actual target data, rather than artificial scaling artifacts.

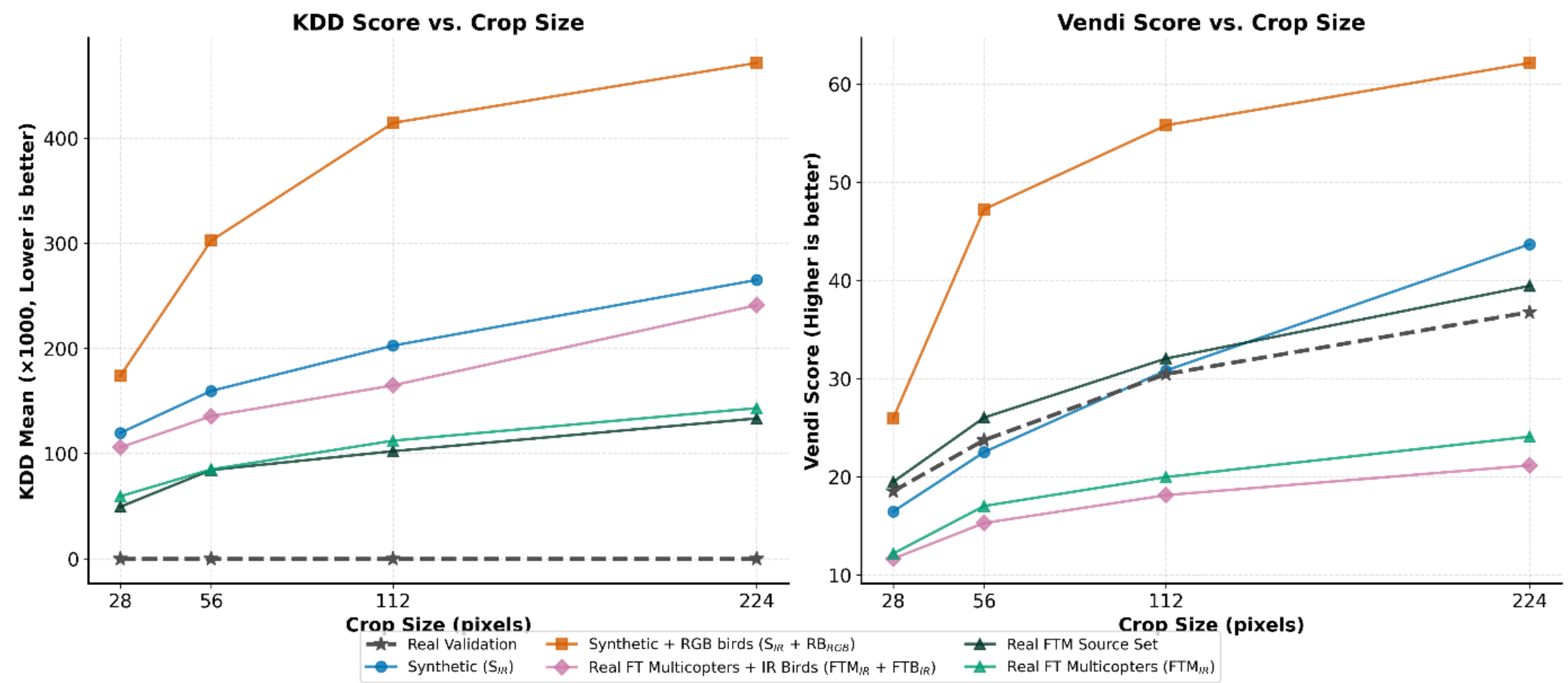

Figure 6. Impact of crop feature extraction resolution on domain gap (KDD) and diversity (Vendi).

The 3-channel input images (R=G=B) were converted to PyTorch tensors and normalized using standard ImageNet mean ([0.485, 0.456, 0.406]) and standard deviation ([0.229, 0.224, 0.225]) to align with the model's pre-training distribution. Feature embeddings were then extracted directly from the global semantic [CLS] token, which effectively aggregates the overarching semantic context. Finally, prior to computing any metrics, these embeddings were L2-normalized, to ensure the evaluation captures true structural alignment rather than being skewed by raw activation magnitudes.

### C.3 Level II: Representation-Space Alignment (Kernel DINO Distance)

The semantic domain gap between datasets was quantified using Kernel DINO Distance (KDD), an adaptation of Kernel Inception Distance [41]. Unlike Fréchet Inception Distance (FID), which assumes Gaussian distributions and exhibits sample-size bias [64], the MMD-based KDD provides an unbiased estimate. While FID and KID rely on the Inception-V3 network, recent studies demonstrate that those features fail to capture complex semantic similarities and align poorly with human judgements [65]. Following the suggestion of Stein et al. [65] we use DINOv2 to leverage a significantly richer feature space, allowing for a much more accurate evaluation of domain alignment. KDD is formulated as the squared Maximum Mean Discrepancy (MMD) between the feature embeddings of the target dataset and the reference dataset, computed using the standard KID polynomial kernel. KDD was calculated across 500 random subsets of 50 samples (sampled without replacement per subset).

### C.4 Level III: Structural Diversity

Dataset diversity was measured using the Vendi Score [60], which estimates the effective sample size from the entropy of eigenvalues derived from a feature similarity matrix. Because the score depends strongly on the feature representation [65], all calculations were performed using the standardized DINOv2 embeddings and a linear kernel ($k(x, y) = x^T y$).

While raw Vendi scores can be biased by dataset size, often requiring truncation to compare uneven datasets [66], we mitigate this issue by leveraging the dual formulation available in the official Vendi Score implementation (score_dual()). Rather than calculating the traditional sample covariance, this method computes the score using the feature covariance matrix. Consequently, the maximum possible score is strictly capped by the dimensionality of our DINOv2 embeddings (D = 1024 for ViT-L). This mathematical bottleneck significantly reduces the volume bias for any dataset containing more than 1,024 images. While our few-shot fine-tuning sets ($FTM_{IR}$, $FTB_{IR}$) remain mathematically constrained by their low image count, all of our other datasets comfortably exceed this threshold. By utilizing this built-in dimensionality cap, our large-scale sets are placed on a more level playing field, which allows for a more robust comparison of structural diversity, though it does not entirely eliminate the potential influence of sample size or modality bias. To empirically validate that our results are not affected by sample size bias, a comprehensive fixed-volume sensitivity analysis, evaluating stability across downsampled tiers, is detailed in Appendix F.

## APPENDIX D: PIXEL-LEVEL METRIC RESULTS

Figure 7 illustrates mean luminance distributions across datasets. Real LWIR imagery shows the expected comb-like quantization pattern produced by AGC, the hardware algorithm that stretches high-bit-depth raw radiometric measurements (typically 14-bit or 16-bit) into the 8-bit display range. In contrast, synthetic images produce a continuous histogram.

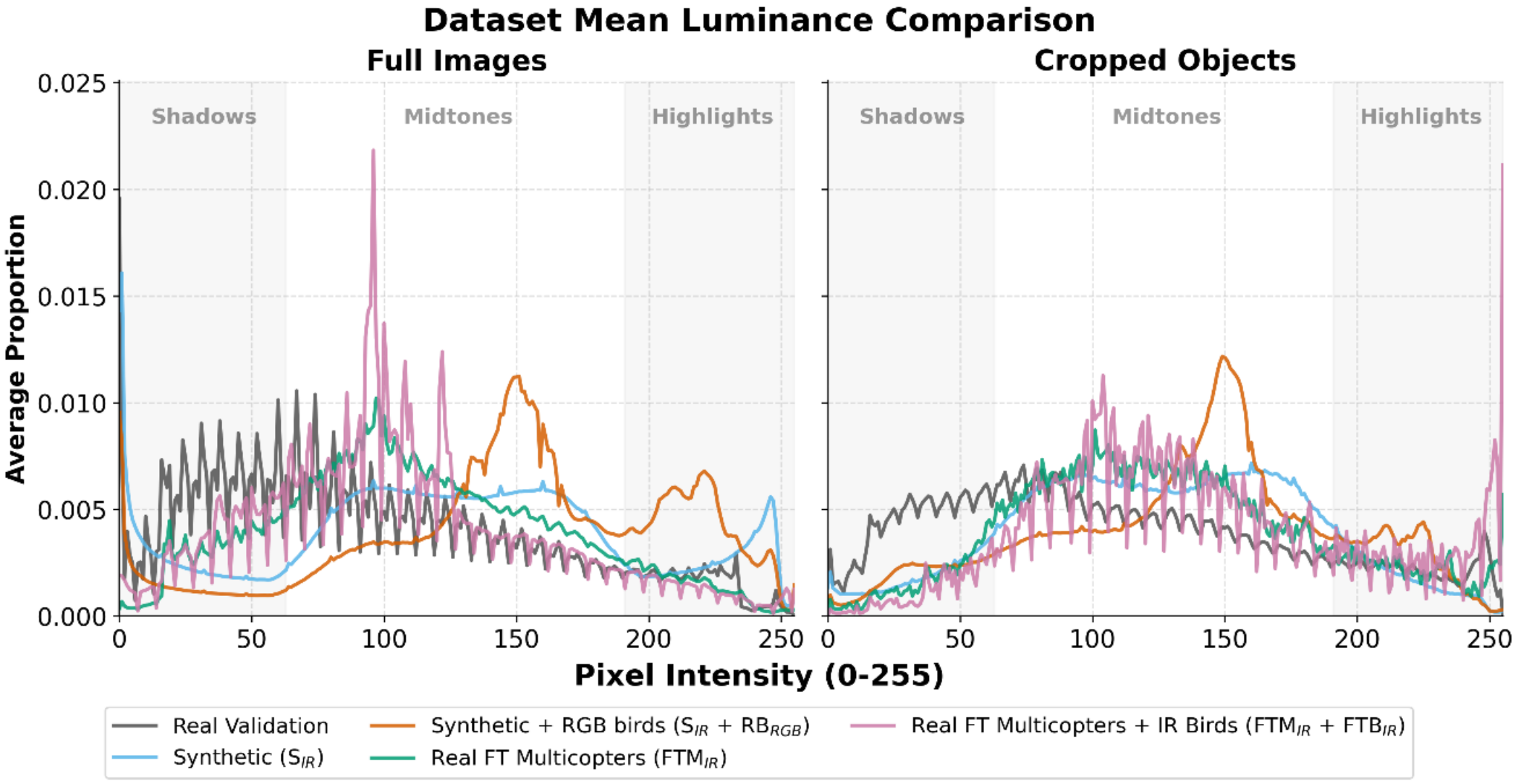

Figure 7. Dataset mean luminance histogram comparison.

Notably, a substantial distributional shift exists even among the physical real-world datasets: the Real Validation set (dark gray line), the $FTM_{IR}$ (green line), and the $FTM_{IR}$ + $FTB_{IR}$ set (pink line) exhibit distinctly different intensity peaks and dynamic range utilizations. The addition of the bird dataset (pink line) is particularly revealing, as it displays highly pronounced quantization spikes. This intra-domain variance highlights a critical challenge in thermal computer vision: physical IR cameras apply proprietary, dynamic AGC stretching logic that fundamentally alters the radiometric profile of the scene based on specific hardware. This likely deepens the thermal data scarcity problem, as a model trained exclusively on one sensor's specific thermal profile may find it harder to generalize to another's.

These differences are confirmed by Earth Mover's Distance (EMD) comparisons, measured against the validation set. The synthetic ($S_{IR}$) dataset alone exhibits an EMD of 29.88. Introducing the RGB bird dataset to the synthetic data (S + $RB_{RGB}$) substantially degrades radiometric alignment, increasing the EMD to 52.80. In comparison, the Real $FTM_{IR}$ dataset demonstrates an EMD of 18.77. Interestingly, the EMD for the Real $FTM_{IR}$ + $FTB_{IR}$ configuration drops to 13.61, indicating that the IR birds dataset has an overall pixel distribution that more closely matches the validation baseline.

Spatial-frequency metrics (Figure 8) further define the systemic physics gap between the domains, with full-image Histogram Total Variation (HTV) yielding the largest statistical divergence between the real validation baseline and the pure synthetic ($S_{IR}$) set (d = 1.242). HTV mathematically captures the severity of the AGC quantization spikes. Because the synthetic pipeline does not replicate this scene-adaptive remapping, it lacks this fundamental quantization roughness. The absence of AGC also causes synthetic environments to exhibit lower overall dynamic range utilization (d = 0.899). Despite these radiometric differences, the application of post-processing noise to the synthetic pipeline closely aligned with the macro-level spatial noise of physical sensors. As shown in Table 4, the median full-image GLCM Correlation (0.994 synthetic vs. 0.977 real validation) and Entropy (9.862 synthetic vs. 9.786 real validation) are highly similar.

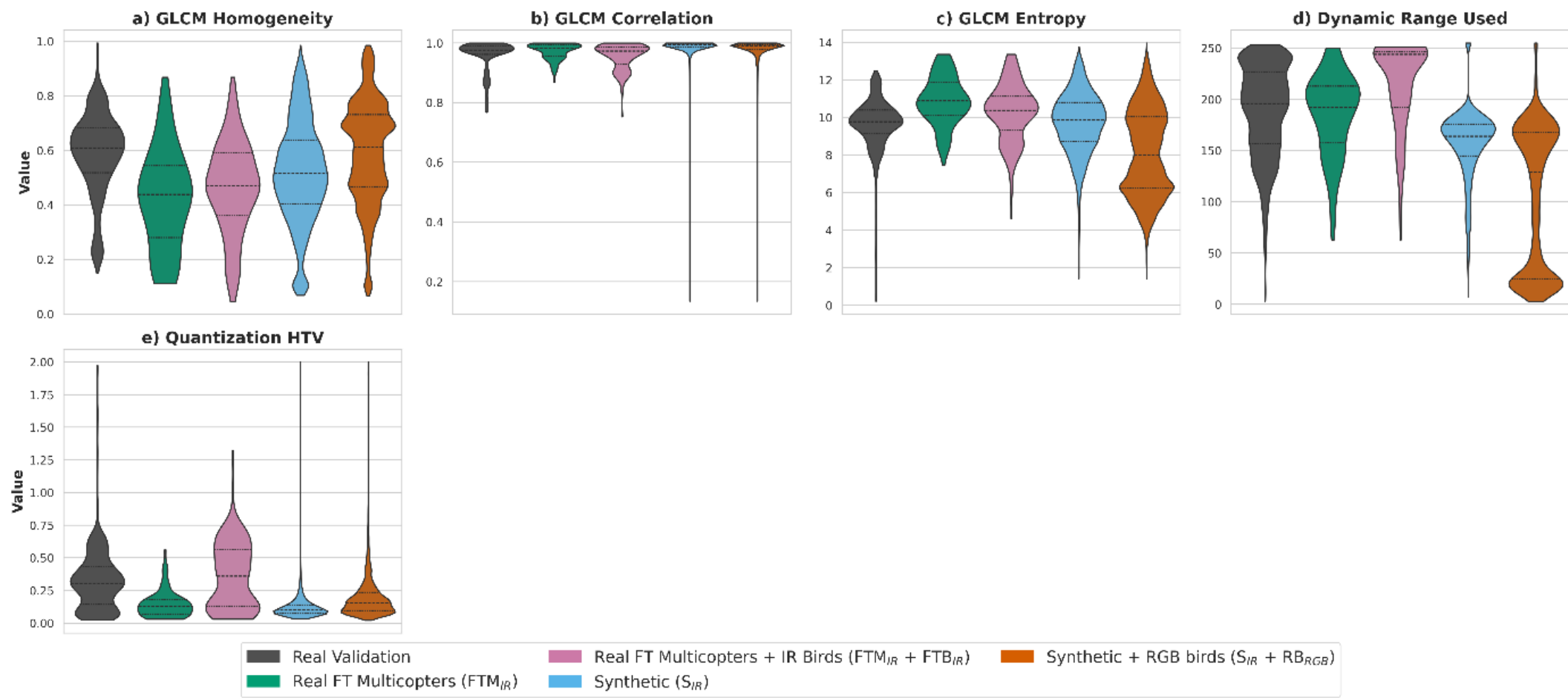

Figure 8. Distributions of Level I metrics across the evaluated datasets. Dashed lines denote the 25th, 50th (median), and 75th percentiles.

Evaluating radiometry at the isolated target level (Figure 9) requires strict statistical guardrails. As established in the methodology, a spatial minimum of 400 pixels was enforced to prevent finite-sample artifacts (bin-starvation) in localized textural calculations. Because the natively small real IR birds fall below this threshold, they were excluded from these specific histogram and GLCM calculations to preserve statistical integrity. Without this filter, there are severe mathematical artifacts due to matrix sparsity, inflating target Quantization HTV and distorting GLCM texture profiles. Conversely, metrics independent of bin-capacity constraints (Local Weber Contrast and Sobel Gradient Variance) do include the infrared birds set.

For the other sets that satisfy this resolution threshold, GLCM texture analysis indicates that real LWIR target crops consistently exhibit internal texture and localized intensity fluctuations. This variation likely reflects a mixture of physical thermal structure, heat diffusion, microbolometer noise, camera-side processing, and compression artifacts. Synthetic targets, in contrast, are rendered as relatively uniform meshes, which explains their lower texture entropy ($d = 0.974$) and different GLCM profiles. Furthermore, this internal uniformity results in a severely lower Sobel Gradient Variance for the synthetic targets ($d = 1.224$), confirming that the simulated drones act as uniform silhouettes rather than noisy, physical heat sources. The complete quantitative breakdown of the pixel-level and texture metrics across all evaluated configurations is provided in Table 4. To isolate the primary drivers of this domain gap, Table 5 highlights only the metrics exhibiting a medium-to-large statistical effect size ($d >= 0.7$) between the real validation data and the pure synthetic data.

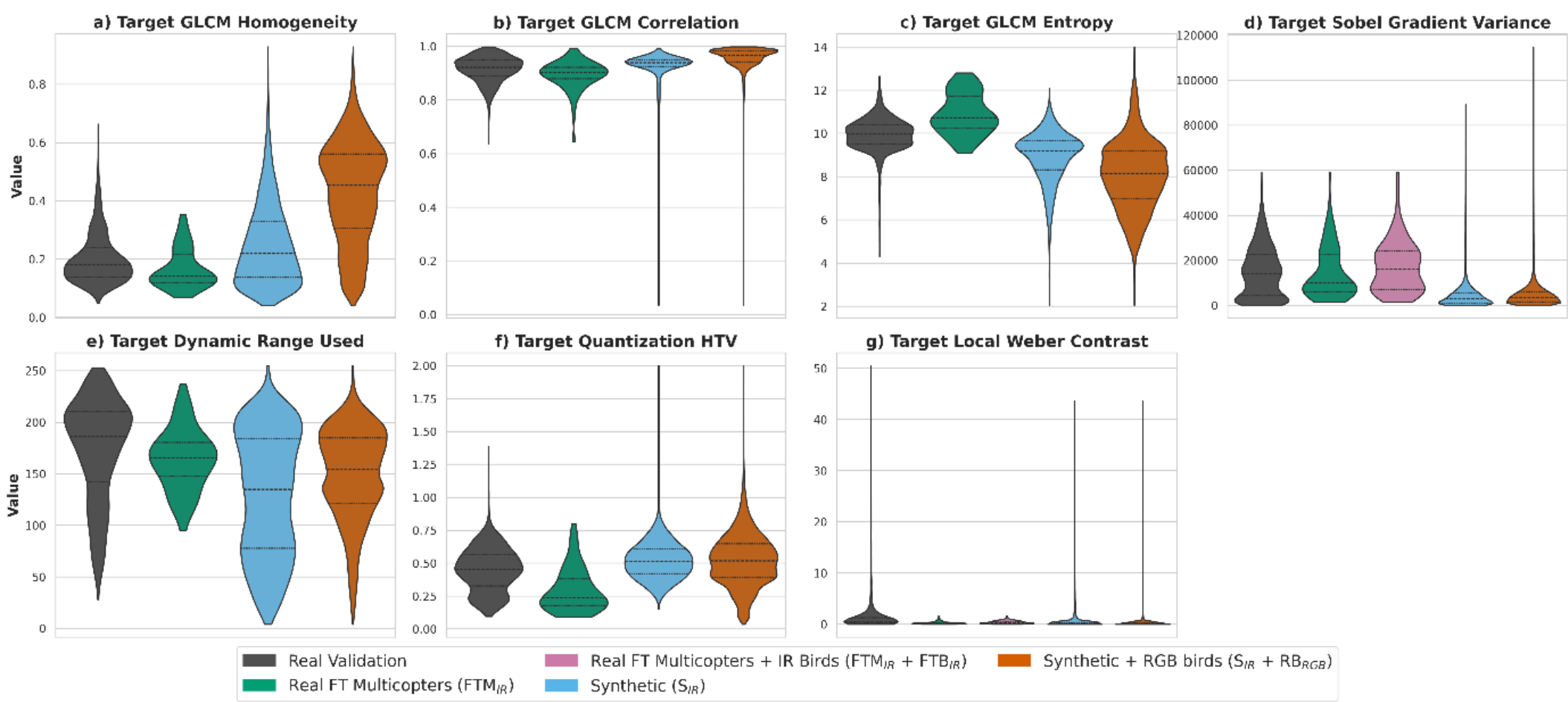

Figure 9. Pixel level metrics for target crops.

Table 4. Full Pixel-Level Metrics (Median (IQR)). Note: Crop histogram and GLCM texture metrics are omitted (—) for the "$FTM_{IR}$ + $FTB_{IR}$" configuration. This is because the bird targets fall below the resolution floor required for valid distributional analysis.

| **Metric** | **Real Validation** | **Synthetic ($S_{IR}$)** | **Synthetic + RGB birds ($S_{IR}$ + $RB_{RGB}$)** | **Real $FTM_{IR}$** | **Real $FTM_{IR}$ + $FTB_{IR}$** |
|---|---|---|---|---|---|
| Full: Mean Luminance | 90.466 (24.210) | 126.575 (59.070) | 148.762 (60.058) | 111.691 (28.788) | 107.429 (27.079) |
| Full: Dynamic Range Utlization | 196.000 (70.000) | 164.000 (31.000) | 129.000 (143.000) | 192.000 (55.250) | 244.000 (54.202) |
| Full: Quantization HTV | 0.306 (0.288) | 0.101 (0.066) | 0.157 (0.140) | 0.131 (0.108) | 0.361 (0.432) |
| Full: GLCM Homogeneity | 0.610 (0.164) | 0.517 (0.235) | 0.613 (0.266) | 0.439 (0.263) | 0.471 (0.230) |
| Full: GLCM Correlation | 0.977 (0.029) | 0.994 (0.010) | 0.989 (0.017) | 0.983 (0.038) | 0.973 (0.057) |
| Full: GLCM Entropy | 9.786 (1.254) | 9.862 (2.089) | 8.000 (3.783) | 10.888 (1.750) | 10.394 (1.800) |
| Crop: Mean Luminance | 105.706 (49.583) | 126.931 (62.791) | 147.973 (53.532) | 124.490 (36.268) | — |
| Crop: Dynamic Range | 186.060 (67.860) | 135.000 (106.150) | 154.360 (63.500) | 165.830 (32.328) | — |
| Crop: Quantization HTV | 0.454 (0.238) | 0.514 (0.188) | 0.519 (0.257) | 0.238 (0.202) | — |
| Crop: GLCM | 0.181 (0.100) | 0.219 (0.192) | 0.454 (0.255) | 0.142 (0.097) | — |

| Homogeneity | | | | | |
|---|---|---|---|---|---|
| Crop: GLCM Correlation | 0.923 (0.060) | 0.940 (0.025) | 0.968 (0.041) | 0.902 (0.043) | — |
| Crop: GLCM Entropy | 9.972 (0.899) | 9.182 (1.352) | 8.150 (2.195) | 10.735 (1.491) | — |
| Crop: Mean Target Contrast | 0.572 (1.049) | 0.194 (0.231) | 0.153 (0.175) | 0.182 (0.200) | 0.336 (0.448) |
| Crop: Sobel Gradient Variance | 14023.100 (18149.501) | 2948.264 (4697.725) | 3595.395 (4486.979) | 10015.921 (16532.323) | 16112.491 (17193.841) |

Table 5. Effect Size (Cohen's d) - Real Validation vs. Synthetic ($S_{IR}$) Base. (Note: Table isolates metrics with d >=0.7. Following standard conventions, d >= 0.8 constitutes a large effect).

| **Metric** | **Cohen's d** |
|---|---|
| Full: Quantization HTV | 1.242 |
| Crop: Sobel Gradient Variance | 1.224 |
| Crop: GLCM Entropy | 0.974 |
| Full: Dynamic Range Used | 0.899 |
| Crop: Dynamic Range Used | 0.763 |

## APPENDIX E: SEMANTIC GAP AND DATASET DIVERSITY RESULTS

To evaluate feature-space alignment, we computed Kernel DINO Distance (KDD) and Vendi Score for both full frames and object crops (Table 6). All KDD values are scaled by 1000 for readability.

Table 6. Domain Gap (KDD) and Diversity (Vendi) analysis results. Higher Vendi indicates greater diversity; desirable values depend on alignment with the target domain. KDD – Lower is better.

| **Dataset** | **Full Image Vendi Score** | **Full Image KDD (x 1000)** | **Cropped Vendi Score** | **Cropped KDD (x 1000)** |
|---|---|---|---|---|
| Validation | 39.5535 | 0 (reference) | 23.7231 | 0 (reference) |
| Real FTM Source Set | 32.2847 | 262.29 ± 39.85 | 26.0178 | 84.33 ± 13.68 |
| Synthetic ($S_{IR}$) | 134.6365 | 332.39 ± 41.95 | 22.5073 | 159.53 ± 20.18 |
| $S_{IR}$ + $RB_{RGB}$ | 100.193 | 367.62 ± 49.66 | 47.2121 | 302.45 ± 53.45 |
| $FTM_{IR}$ | 27.6318 | 222.68 ± 31.43 | 17.0193 | 84.91 ± 12.37 |
| $FTM_{IR}$ + $FTB_{IR}$ | 29.683 | 272.79 ± 42.77 | 15.2835 | 135.56 ± 20.84 |

## APPENDIX F: VENDI SCORE SENSITIVITY ANALYSIS

To validate that the dual formulation successfully isolates underlying structural diversity from residual volumetric advantages, we conducted a multi-tier fixed-volume sensitivity analysis across three down sampled ceilings ($N \in$ \\\\{1000,500,100\\\\}).

For the large-scale datasets, random subsampling was performed across three distinct evaluation seeds and averaged to capture statistical variance. For the $N = 100$ tier, our 100-sample fine-tuning ($FTM_{IR}$) dataset was integrated directly, as its native volume perfectly matches this evaluation ceiling. By strictly equalizing evaluation volumes, this ablation provides strong evidence that sample size is not the primary driver of the observed variance in Vendi scores. As detailed in Table 7, the relative dataset hierarchy remains stable across all three down sampled tiers.

Table 7. Fixed-Volume Sensitivity Analysis of Vendi Scores. Structural diversity evaluated across three uniform downsampled tiers.

| **Evaluation Target & Dataset** | **Baseline (Full Set)** | **Matched N=1000** | **Matched N=500** | **Matched N=100** |
|---|---|---|---|---|
| Full-Frame Diversity (Vendi) | | | | |
| Synthetic ($S_{IR}$) ($N = 20{,}000$) | 134.64 | 111.20 (± 2.85) | 93.87 (± 1.29) | 47.32 (± 2.29) |
| Real Validation ($N = 3{,}000$) | 39.55 | 36.42 (± 0.38) | 33.42 (± 0.44) | 21.83 (± 0.84) |
| Real FTM Source Set ($N = 2{,}700$) | 32.28 | 30.34 (± 0.28) | 28.48 (± 0.91) | 20.82 (± 0.69) |
| $FTM_{IR}$ ($N = 100$) | 27.63 | — | — | 27.63 (Native) |
| Target Crop Diversity (Vendi) | | | | |

| Synthetic ($S_{IR}$) ($N = 20{,}000$) | 22.51 | 20.99 (± 0.35) | 19.33 (± 0.66) | 13.46 (± 0.79) |
|---|---|---|---|---|
| Real Validation ($N = 3{,}000$) | 23.72 | 21.99 (± 0.33) | 19.90 (± 0.08) | 12.71 (± 0.57) |
| Real FTM Source Set ($N = 2{,}700$) | 26.02 | 24.31 (± 0.33) | 21.48 (± 0.28) | 14.50 (± 0.44) |
| $FTM_{IR}$ ($N = 100$) | 17.02 | — | — | 17.02 (Native) |

At the full-frame level, the relative mathematical distance and ordering of the randomly sampled datasets are preserved precisely from the un-subsampled baseline down to the $N = 100$ ceiling. Similarly, the localized target (crop) hierarchy consistently inverts and maintains its exact structural sequence across all matched volumes. While absolute scores naturally compress as the sample ceiling lowers, the preservation of the relative bounds demonstrates that the underlying feature distributions are stable.

Additionally, the $N = 100$ tier reveals a distinct diversity delta between the randomly subsampled pools and the natively 100-sample fine-tuning set ($FTM_{IR}$). The dataset sampled via K-Means clustering exhibits a measurable diversity advantage over its randomly sampled counterpart (e.g., native $FTM_{IR}$ $N = 100$ full images at 27.63 versus the random $N = 100$ drawing at 20.82). This divergence verifies that the sampling protocol successfully optimizes feature space coverage at low volumes.

Because strict volume-matching preserves the comparative dataset hierarchy across all randomly subsampled datasets, we empirically demonstrate that the $D = 1024$ dimensionality cap effectively controls for sample size effects—the only exception being the intentionally selected native 100-sample subset ($FTM_{IR}$), as it was optimized for feature-space coverage. This validates the mathematical fairness and reliability of utilizing the full-scale dual Vendi scores in our primary evaluations.

## APPENDIX G: MODEL PERFORMANCE

Tanle 8 provides an overview of the model performance across training protocols and models. Except for a few cases of fine-tuning, the overall improvements in recall, precision, mAP@50 and mAP@50:95 shown in Table 8 across all models demonstrate that the fine-tuning strategies adopted in this study are model-agnostic. This thorough evaluation confirms the quality of the IR-generated synthetic scenes and highlights its potential as a high-performance pre-training resource for the G2A drone detection task in the LWIR domain.

Table 8. Model performance. $S_{IR}$: pre-training on Synthetic data; $S_{IR} + RB_{RGB}$: pre-training on Synthetic and proprietary RGB bird data; $(S_{IR})_w \rightarrow FTM_{IR}$ : weights were initialized from pre-training on Synthetic data, then Fine-Tuned on real-world LWIR multicopter data; $(S_{IR} + RB_{RGB})_w \rightarrow FTMB_{IR}$: weights were initialized from pre-training on Synthetic and proprietary RGB bird data, then Fine-Tuned on real-world LWIR multicopter and bird data; $RM_{IR}$: no pre-training was performed. The models were trained solely on Real-World IR multicopter data.

| **Model Size** | **Training Set and Size** | **Model** | **Recall** | **Precision** | **mAP@50** | **mAP@50:95** |
|---|---|---|---|---|---|---|
| n | $S_{IR}$ | YOLOv13 | 0.698 | 0.910 | 0.769 | 0.409 |
| | [20 K] | RF-DETR | 0.802 | 0.926 | 0.857 | 0.437 |
| | $S_{IR} + RB_{RGB}$ | YOLOv13 | 0.707 | 0.901 | 0.797 | 0.518 |
| | [39,875] | RF-DETR | 0.770 | 0.907 | 0.844 | 0.495 |
| | $(S_{IR})_{w \rightarrow} FTM_{IR}$ | YOLOv13 | 0.791 | 0.928 | 0.883 | 0.610 |
| | [100] | RF-DETR | 0.930 | 0.978 | 0.961 | 0.649 |
| | $(S_{IR} + MB_{RGB})_w \rightarrow$ | YOLOv13 | 0.760 | 0.885 | 0.850 | 0.604 |
| | $FTMB_{IR}$ [200] | RF-DETR | 0.900 | 0.950 | 0.948 | 0.651 |
| | $RM_{IR}$ | YOLOv13 | 0.767 | 0.893 | 0.847 | 0.567 |
| | [100] | RF-DETR | 0.910 | 0.955 | 0.942 | 0.620 |
| s | $S_{IR}$ | YOLOv13 | 0.564 | 0.829 | 0.670 | 0.378 |
| | [20 K] | RF-DETR | 0.835 | 0.939 | 0.893 | 0.474 |
| | $S_{IR} + RB_{RGB}$ | YOLOv13 | 0.617 | 0.812 | 0.704 | 0.442 |
| | [39,875] | RF-DETR | 0.792 | 0.886 | 0.873 | 0.538 |
| | $(S_{IR})_{w \rightarrow} FTM_{IR}$ | YOLOv13 | 0.834 | 0.910 | 0.890 | 0.612 |
| | [100] | RF-DETR | 0.950 | 0.975 | 0.974 | 0.689 |
| | $(S_{IR} + MB_{RGB})_w \rightarrow$ | YOLOv13 | 0.787 | 0.948 | 0.872 | 0.634 |
| | $FTMB_{IR}$ [200] | RF-DETR | 0.950 | 0.973 | 0.967 | 0.692 |
| | $RM_{IR}$ | YOLOv13 | 0.765 | 0.912 | 0.855 | 0.577 |
| | [100] | RF-DETR | 0.940 | 0.971 | 0.971 | 0.670 |
| m | $S_{IR}$ | RF-DETR | 0.796 | 0.921 | 0.864 | 0.467 |
| | [20 K] | | | | | |
| | $S_{IR}$ + $RB_{RGB}$ [39,875] | RF-DETR | 0.771 | 0.955 | 0.846 | 0.501 |
| | $(S_{IR})_w \rightarrow FTM_{IR}$ [100] | RF-DETR | 0.960 | 0.984 | 0.980 | 0.705 |
| | $(S_{IR} + MB_{RGB})_w \rightarrow FTMB_{IR}$ [200] | RF-DETR | 0.940 | 0.977 | 0.975 | 0.719 |
| | $RM_{IR}$ [100] | RF-DETR | 0.930 | 0.985 | 0.969 | 0.663 |
| l | $S_{IR}$ [20 K] | YOLOv13 | 0.632 | 0.856 | 0.734 | 0.382 |
| | | RF-DETR | 0.796 | 0.934 | 0.880 | 0.463 |
| | $S_{IR}$ + $RB_{RGB}$ [39,875] | YOLOv13 | 0.660 | 0.884 | 0.741 | 0.451 |
| | | RF-DETR | 0.728 | 0.934 | 0.839 | 0.522 |
| | $(S_{IR})_w \rightarrow FTM_{IR}$ [100] | YOLOv13 | 0.819 | 0.945 | 0.897 | 0.656 |
| | | RF-DETR | 0.950 | 0.974 | 0.976 | 0.713 |
| | $(S_{IR} + MB_{RGB})_w \rightarrow FTMB_{IR}$ [200] | YOLOv13 | 0.760 | 0.920 | 0.870 | 0.615 |
| | | RF-DETR | 0.960 | 0.978 | 0.984 | 0.723 |
| | $RM_{IR}$ [100] | YOLOv13 | 0.724 | 0.922 | 0.834 | 0.593 |
| | | RF-DETR | 0.930 | 0.965 | 0.972 | 0.698 |
| x | $S_{IR}$ [20 K] | YOLOv13 | 0.644 | 0.840 | 0.733 | 0.386 |
| | $S_{IR}$ + $RB_{RGB}$ [39,875] | YOLOv13 | 0.610 | 0.897 | 0.721 | 0.430 |
| | $(S_{IR})_w \rightarrow FTM_{IR}$ [100] | YOLOv13 | 0.787 | 0.942 | 0.885 | 0.637 |
| | $(S_{IR} + MB_{RGB})_w \rightarrow FTMB_{IR}$ [200] | YOLOv13 | 0.787 | 0.924 | 0.876 | 0.622 |
| | $R M_{IR}$ [100] | YOLOv13 | 0.760 | 0.947 | 0.863 | 0.616 |
| | $S_{IR}$ [20 K] | Faster-RCNN | 0.424 | 0.581 | 0.581 | 0.287 |
| | $S_{IR}$ + $RB_{RGB}$ [39,875] | Faster-RCNN | 0.443 | 0.658 | 0.658 | 0.331 |
| | $(S_{IR})_w \rightarrow FTM_{IR}$ [100] | Faster-RCNN | 0.488 | 0.672 | 0.672 | 0.364 |
| | $(S_{IR} + MB_{RGB})_w \rightarrow FTMB_{IR}$ [200] | Faster-RCNN | 0.518 | 0.511 | 0.511 | 0.290 |
| | $RM_{IR}$ [100] | Faster-RCNN | 0.348 | 0.035 | 0.035 | 0.012 |

## APPENDIX H: MODEL PARAMETER COUNT, COMPUTATIONAL COMPLEXITY AND INFERENCE SPEED COMPARISON

This appendix presents the computational complexity data and inference latency benchmarking results for the evaluated object detection frameworks. The comparative metrics detailing models' mAP@50:95 against parameter count, GFLOPs, latency, and frame rates across all training configurations are compiled in Figure 10 below. Inference latencies were benchmarked on an NVIDIA T4 GPU utilizing TensorRT 11 (FP16 precision) via the NVIDIA *trtexec* utility, averaged over 100 runs following an initial GPU warm-up phase. Due to structural incompatibilities between Faster R-CNN's dynamic, data-dependent Region of Interest (RoI) pooling layers and TensorRT's strict constraints, Faster R-CNN could not be converted to FP16. Consequently, its latency was benchmarked natively in PyTorch (FP32 precision).

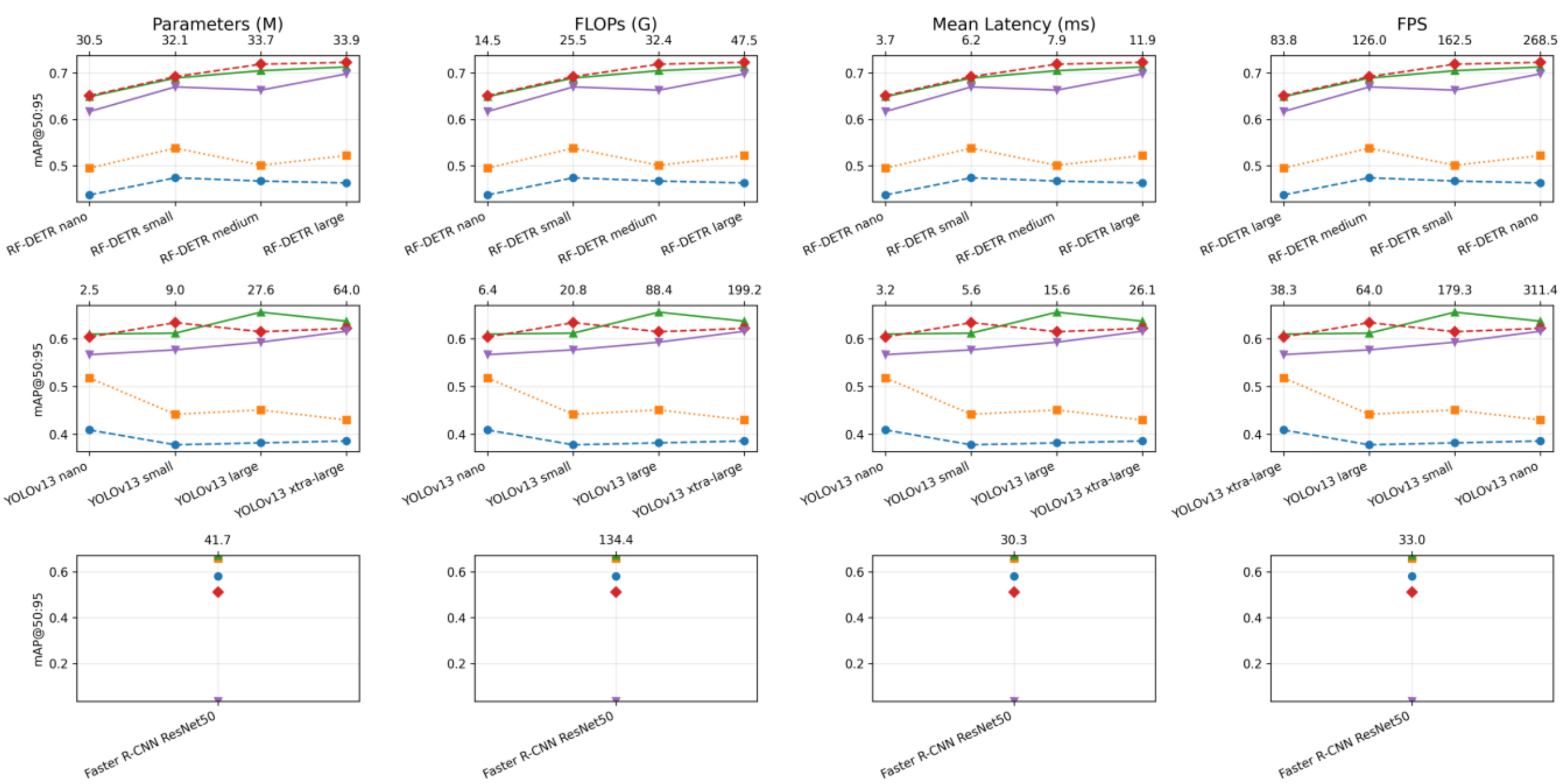

Figure 10: Models' mAP@50:95 against parameter count, GFLOPs, latency, and frame rates across all training configurations on an NVIDIA T4 GPU.